\documentclass[iicol,referee]{sn-jnl}

\usepackage{natbib}
\usepackage{hyperref}
\usepackage{comment}
\hypersetup{colorlinks=true,linkcolor=blue,urlcolor=blue}
\jyear{2022}%

\theoremstyle{thmstyleone}%
\newtheorem{theorem}{Theorem}
\newtheorem{proposition}[theorem]{Proposition}%

\theoremstyle{thmstyletwo}%
\newtheorem{example}{Example}%
\newtheorem{remark}{Remark}%
\usepackage{comment}
\theoremstyle{thmstylethree}%
\newtheorem{definition}{Definition}%

\begin{document}

\title[Article Title]{Online Handwriting Trajectory Reconstruction from Kinematic Sensors using Temporal Convolutional Network}


\author*[1]{\fnm{Wassim} \sur{Swaileh}}\email{wassim.swaileh@irisa.fr}
\equalcont{These authors contributed equally to this work.}

\author*[1]{\fnm{Florent} \sur{Imbert}}\email{florent.imbert@irisa.fr}
\equalcont{These authors contributed equally to this work.}

\author[2]{\fnm{Yann} \sur{Soullard}}

\author[2]{\fnm{Romain} \sur{Tavenard}}

\author[1]{\fnm{Eric} \sur{Anquetil}}


\affil[1]{\orgname{IRISA, Université de Rennes, INSA Rennes}, \orgaddress{
 \country{France}}}

\affil[2]{\orgdiv{LETG}, \orgname{IRISA, Université Rennes 2}, \orgaddress{35043 Rennes, \country{France}}}

\newcommand{\yann}{\textcolor{black}}
\newcommand{\romain}{\textcolor{black}}
\newcommand{\florent}{\textcolor{black}}
\newcommand{\Florent}{\textcolor{black}}
\newcommand{\wassim}{\textcolor{black}}

\vspace{-2cm}
\abstract{

\textcolor{blue}{\romain{Handwriting with digital pens is a common way to facilitate human-computer interaction through the use of Online Handwriting (OH) trajectory reconstruction. In this work, we focus on a digital pen equipped with sensors from which one wants to reconstruct the OH trajectory. Such a pen allows to write on any surface and to get the digital trace, which can help learning to write, by writing on paper, and can be useful for many other applications such as collaborative meetings, etc. In this paper, we introduce a novel processing pipeline that maps the sensor signals of the pen to the corresponding OH trajectory. Notably, in order to tackle the difference of sampling rates between the pen and the tablet (which provides ground truth information), our preprocessing pipeline relies on Dynamic Time Warping to align the signals. We introduce a dedicated neural network architecture, inspired by a Temporal Convolutional Network, to reconstruct the online trajectory from the pen sensor signals. Finally, we also present a new benchmark dataset on which our method is evaluated both qualitatively and quantitatively, showing a notable improvement over its most notable competitor.}}
}




\maketitle

\section{Introduction}\label{sec1}
Digital devices can help pupils and teachers in the learning process by promoting active learning techniques and providing immediate feedbacks~\cite{simonnet2019evaluation}. The e-learning literature shows that computer-based analysis of handwriting can be really accurate, sensitive, and reliable to produce relevant and consistent feedbacks for correction or guidance. Several pen-based tablet applications have been designed in order to give immediate and personalized feedback to children \cite{krichen2022combination}. 
Moreover, children still need to learn to write on paper to acquire different sensations because today, it is still by far the most widely used handwriting medium.  

As an answer to that need, digital pens have been designed to allow for handwriting on paper while capturing the handwriting gesture. Here, we focus on such kinds of stylus with the goal to reconstructing the digital handwriting trajectory of the pen. The digital pen that we use in this work is the Digipen stylus developed by STABILO, which is equipped of kinematic sensors to track the pen movements. 


Nowadays, a wide range of applications in the domain of remote sensing and tracking systems benefits from recent improvements in deep neural network architectures. 
\yann{Tracking systems are commonly based on Inertial Measurement Unit (IMU) due to the low cost of these sensors. However, IMU-based sensors are quite imprecise due to the poor quality of the IMU signals.} 


\wassim{IMU sensors are utilized in other fields to recognize pre-defined movements \cite{bracelet,ring} or recreate pedestrian trajectories \cite{Shoes}.}
\textcolor{blue}{\romain{More closely related to this work, a wide range of studies have tackled the OH recognition task, which can be successfully accomplished despite the noisy nature of IMU sensor signals.}} 

In this work, we focus on the more challenging task of trajectory reconstruction.
\yann{Regarding OH recognition, there is one label for a global shape of handwriting. The model is trained to extract global displacement features to produce a prediction. This is in contrast with the OH reconstruction where there is one label per time frame. Thus, the model is trained to extract local displacement features based on a more or less local view to produce position predictions. In addition, hovering parts, i.e. when the pen is in air, have to be modeled in order to get a good position of the next touching stroke. This must be done without knowledge about the labeling, as there is no pen traces related to those parts.} 
\yann{To the best of our knowledge, the recent work proposed by Wehbi et al. \cite{wehbi2022surface} is the second attempt to reconstruct OH trajectories from a digital stylus using deep neural networks. It generalizes the works from Ott et al. \cite{ott2022joint} to multiple writers. Older works explore more traditional approaches for this task such as Markovian models \cite{pan2018handwriting} or movement acceleration inference techniques \cite{bu2021handwriting}. 
}

\textcolor{blue}{\romain{In terms of evaluation, some of the earlier works relied on a qualitative assessment of the reconstructed trajectories \cite{nguyen2021online}. In other works, recognition performance is used as a proxy to evaluate reconstruction \cite{huang2022agtgan,wehbi2022surface}. 
Sometimes, additional reconstruction-only metrics such as the Root Mean Squared Error (RMSE) or the Dynamic Time Warping (DTW) are used \cite{chen2022complex}.
}}  

This work presents a novel handwriting trajectory reconstruction pipeline from IMU sensors.  
Our main contributions are: 
\begin{itemize}
    
    \item preprocessing including an alignment strategy between input and ground-truth time series for training, as well as a learning pipeline based on touching strokes,
    \item a neural network architecture inspired by Temporal Convolutional Networks (TCN) to cope with noisy signals from the inertial sensors,
    \item a database that will serve as a benchmark to help advance research, 
    \item an evaluation protocol based on the Fréchet distance to assess the quality of the reconstructed trajectories.
\end{itemize}
Both training and testing phases are described in details to allow the reproducibility of experiments. 

The rest of the paper is organized as follows, related works are presented in the following section. Section 3 introduces the data acquisition protocol and challenges of this task. The proposed method is described in section 4. The experimental results are presented and discussed in section 5 before the conclusion and perspectives. 

\section{Related works}

\yann{While the field of digital devices for note-taking, drawing, or handwriting learning is growing quickly, most of the systems use a screen for digital handwriting acquisition. Only few of them use a stylus that integrates motion tracking systems in order to reconstruct the handwriting trajectories. To the next, we first present related works on handwriting trajectory reconstruction, based on various devices. Then, we focus on the use of IMU sensors for handwriting recognition which is a task that have been widely studied. }

\subsection{Handwriting trajectory reconstruction}
There are several types of systems dedicated to acquisition of digital handwriting. The first one is based on screens with special pen-based tablet. Tablet manufacturers, based on different Wacom technology, have developed their devices -- such as Samsung Spen, Apple Pencil and Microsoft Stylet Surface Pen -- to write on a tablet and enable to have a digital OH. These pen based tablets provide a precise track, but they have to be used on a specific device. 

Another approach is to allow writing on paper, which is more natural, easier and more ergonomic. To do that, pens are equipped with specific tools to capture the handwriting gestures. One way is to embed a camera in the stylus, as for the Anoto. Disadvantages are that these stylus are expensive and that they have to be used on special papers.

A second way is to equip the stylus with IMU sensors.
This solution is surface free and low cost and one can write on tablet, paper, or on a board. However, in contrast to a pen based tablet, where one obtains absolute coordinates of the online trajectory, a signal from IMU sensors is only composed of relative pen displacements and it can be very noisy. 

\romain{Based on the previous statement, we can distinguish between three categories of OH trajectory reconstruction tasks: i) trajectory reconstruction from offline handwriting image \cite{chen2022complex}; ii) reconstruction from a pen-tip optical tracking system \cite{ott2022joint}; iii) trajectory reconstruction from IMU signals \cite{bu2021handwriting, ott2022joint, wehbi2022surface}. 
In practice, only a limited number of works have focused on OH trajectory reconstruction from IMU signals. The works of Wehbi et al., \cite{wehbi2022surface} (multi-writer approach) and Ott et al, \cite{ott2022joint} (mono-writer non-generic approach) used deep learning techniques to reconstruct the OH trajectories from IMU signals as a step towards the OH recognition task. 
Earlier works such as \cite{7574685} used more traditional approaches (Hidden Markov models \cite{pan2018handwriting}, pen-tip acceleration and angular velocity inference to multi-level writing plans \cite{bu2021handwriting}) for the same task.}

\florent{Bu et al. \cite{bu2021handwriting} introduces an IMU system based on accelerometer and gyroscope data to reconstruct handwriting. 
The method infers the tip trace displacements (captured by the IMU) into several planes of OH trajectories using principal component analysis (PCA) and a set of empirically chosen parameters. 
This method works in optimal conditions of use but remains sensitive to rotations and translations of the pen.} 

\textcolor{blue}{\romain{Ott et al. \cite{ott2022joint} rely on multi-task learning for joint classification and trajectory reconstruction. They show that a joint learning improved both the OH recognition and the trajectory reconstruction when using a suitable combination of loss functions. However, the proposed approach cannot be generalized since the proposed architecture is limited to produce trajectories of fixed length (100 points). Furthermore, the training and test sets were collected over a single writer. }}

Liu et al. \cite{MagHacker} focus on magnetic signals but they are sensitive to outer
magnetic field.
In \cite{pan2018handwriting}, the authors use low cost IMU from a smartphone to recognize characters and perform trajectory reconstruction. Linear Discriminant Analysis (LDA) is used to detect movements which are estimated mathematically. 

Recently, Wehbi et al. \cite{wehbi2022surface} have proposed an approach that deals with the Stabilo Digipen, on its earlier version 5.0.
They create a first dataset by using the Digipen on a tablet via a Stabilo mobile application. This  allows to acquire simultaneously input signals from the pen and OH trajectories from a tablet. 

\yann{The gap between sampling rates coming from the pen and the tablet makes mandatory to align the two time series to get one label for each time frame. It is done through a preprocessing that will be discuss more in details in Section~\ref{sec:challenges}.} 
\yann{To do so, Wehbi et al. \cite{wehbi2022surface} linearly interpolate the OH trajectory signal from the tablet. However, the linear interpolation may re-distribute the points equally in time, which result in the loss of the online writing dynamics of speeds and acceleration. }

The authors observe that the linear interpolation on the whole output signal leads to erroneous alignments due to the variability of the transmission time delay. By applying the linear interpolation on strokes rather than on the whole label, better alignments are obtained at the cost of dropping out from the dataset any sample whose input and output signals have unequal numbers of strokes. The number of strokes is deduced from the force sensor signal (pen side) and the pressure signal (tablet side).  Then, they use a Convolutional Neural Network (CNN) made of 3 convolutional layers with batch normalization layers in between to reconstruct the handwriting trajectory. 


\subsection{Using IMU sensors for handwriting recognition}
\romain{As discussed before, only few works perform handwriting reconstruction from IMU sensors. 
On the other hand, the task of OH recognition from IMU sensor data has been tackled through the use of:} \yann{ i) the pen-tip trajectory signal given by a touchscreen device \cite{mustafid2022iamonsense}; ii) signals coming from IMU sensors \cite{bu2021handwriting, ott2022benchmarking, ott2022joint, wehbi2021towards, wehbi2022surface}.}

\yann{Recently, \cite{ott2022benchmarking} introduced a benchmark study that compares several  neural networks architectures trained to recognize characters, symbols, words and equations from IMU signals. A CNN/BLSTM architecture proposed in  \cite{ott2020onhw} obtains the best results. 
Other works intend to address both OH trajectory reconstruction and character recognition. For instance, Wehbi et al. trained two neural networks in succession: first, a model for reconstructing the OH trajectories from IMU signals; second, a model for character recognition based on the reconstructed trajectories. They assess the quality of the reconstruction based on the character recognition rate, \yann{but a good recognition rate may not reflect the quality of the reconstructed trajectory}. 
A multitask learning is proposed by Ott et al. \cite{ott2022joint} using a CNN-LSTM network. They show that both the OH trajectory reconstruction and the character classification benefit from a joint learning. }

\yann{As part of the UbiComp 2021 Challenge on OH recognition, Wegmeth et al. \cite{wegmeth2021detecting} trained a CNN/BLSTM to recognize mathematical expressions written with the Stabilo’s Digipen. Their approach is based on a boundary feature extractor follow by a character classifier. As a consequence, the recognition performance depends on the quality of the label boundary splitting. Other studies minimize the link bandwidth between the Digipen and the remote device (e.g. tablet) \cite{kress2022hardware} or explore domain adaptation \cite{klass2022uncertainty, ott2022domain} and explainability \cite{azimi2022improving} in the context of OH recognition from the Digipen.}
\begin{figure}[!b]
\begin{center}
\includegraphics[scale=0.06]{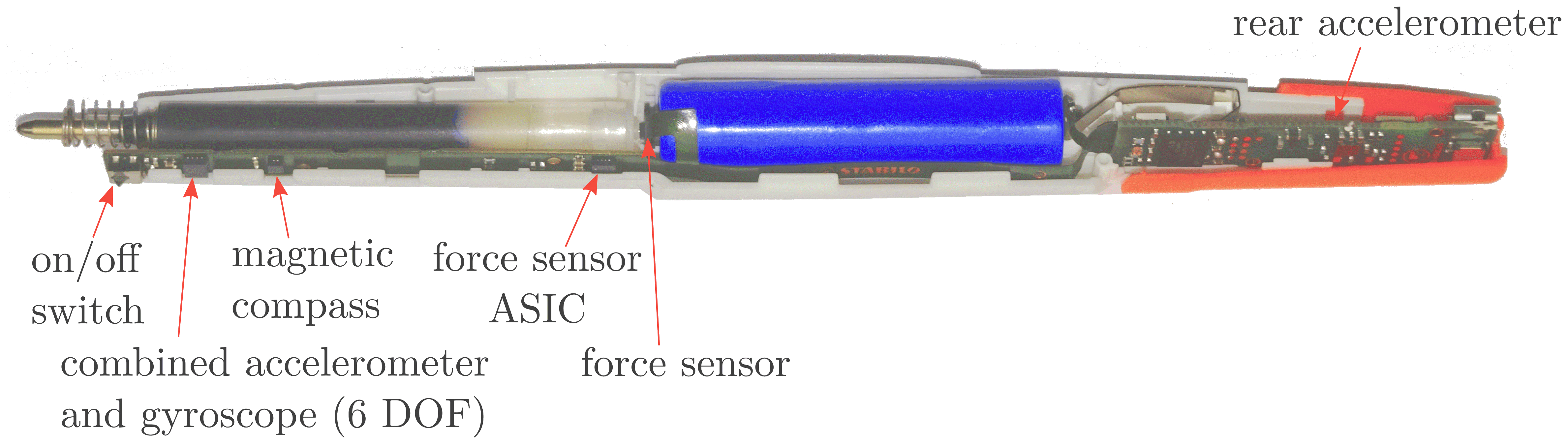}
\caption{© STABILO International Digipen's sensor location }
\label{fig:Sensor_location}
\end{center}
\end{figure}

\section[sec:data acq]{Data acquisition}
\yann{In this section, we present the protocol applied for collecting a new benchmark IRISA-KIHT dataset. We detail the data acquisition process and the protocol to generate the ground truth. } 
\wassim{For this, Stabilo \footnote{https://stabilodigital.com/} created the Digipen, a pen ball digital pen with a Wacom tip instead of an ink ball.}
Several versions of such digital pen exist, such as Digipen kids version 5.0 that works at 100Hz sampling rates and Digipen basic version 6.0 that works at 400Hz (the one used in this work). 

\subsection{Acquisition tools description} \label{sec:acq}
To create the training and test datasets of the handwriting trajectory reconstruction task, we use three equipments and tools: (i) the Digipen; (ii) a Wacom tablet operated with android OS; (iii) the Stabilo's application.

The Digipen embeds the following IMU sensors (Figure \ref{fig:Sensor_location}): a gyroscope, a front and a rear accelerometer, a magnetometer, and a force sensor.
Each of these sensors has its own internal clocking system and provides temporal reading values that describe the relative pen movement in 3-axes channels (x, y and z), except the force sensor which has only one channel.
The readings of the sensors are buffered before being sent to the tablet via Bluetooth connection. 
This makes 13 multi-variate time series for every dataset sample.


     
The tablet choice is important, as the model, the screen size, and the sampling rate must be considered to get consistent OH signals as ground truth. \Florent{Indeed, each tablet has its own sampling rate so, only one model of tablet (Samsumg Galaxy S7 FE) has been kept to obtain homogeneous ground truth. The tablet's base sampling rate is 370Hz which is close to the Digipen one. Note that sampling rate can decrease to 60Hz depending on the writing.} 

Stabilo has developed a dedicated mobile application to record (i) the handwriting trajectories using the Digipen integrating Wacom tip, (ii) the corresponding sensor signals from the Digipen. The application allows to calibrate the Digipen sensors' signals and setup the sampling rate of the Digipen sensors. 
\subsection{Data acquisition protocol}

The recording process (Figure \ref{fig:pipline-training}(a)) starts by selecting one set of predefined scripts that will be written on the tablet surface using the Digipen. \yann{One set consists of 34 samples that have to be written one at a time during a single recording session.
It is composed of five groups: 15 characters, 10 words, 5 equations, 2 shapes and 2 word groups. } 
\yann{While recording, a user holds the pen's on/off switch up, which is a natural way to take the Digipen due to grips designed on the pen to naturally position the fingers properly.} 





\subsection{Data acquisition challenges} \label{sec:challenges}

\yann{To create the IRISA-KIHT dataset the acquisition process is quite challenging due to several difficulties. First, the stylus and the tablet have different sampling rates, so the selection of both of them is an important step. \Florent{As discussed before, we select the Samsung S7 FE tablet and Digipen v6 for data collection.} Second, an important degree of freedom comes from the pen orientation that directly impacts pen sensor values. The Digipen design includes grips to naturally position the fingers properly with the stylus on the same orientation. Another challenge concerns the recording of the hovering (pen up) movements. When the pen gets too far from the tablet, the tablet stops recording a trace. This results to parts of the data coming from the stylus that not match any ground truth from the tablet. }

\wassim{In addition, a drift in the accelerometer measurements is possible, and the kinematic signals are transmitted from the pen to the tablet in sets of six points through Bluetooth. Variable transmission time delay results in asynchronous timestamps for kinematic and tablet data.}

\section{Trajectory reconstruction pipeline}
\begin{figure*}[thpb]
\begin{center}
\includegraphics[scale=0.54]{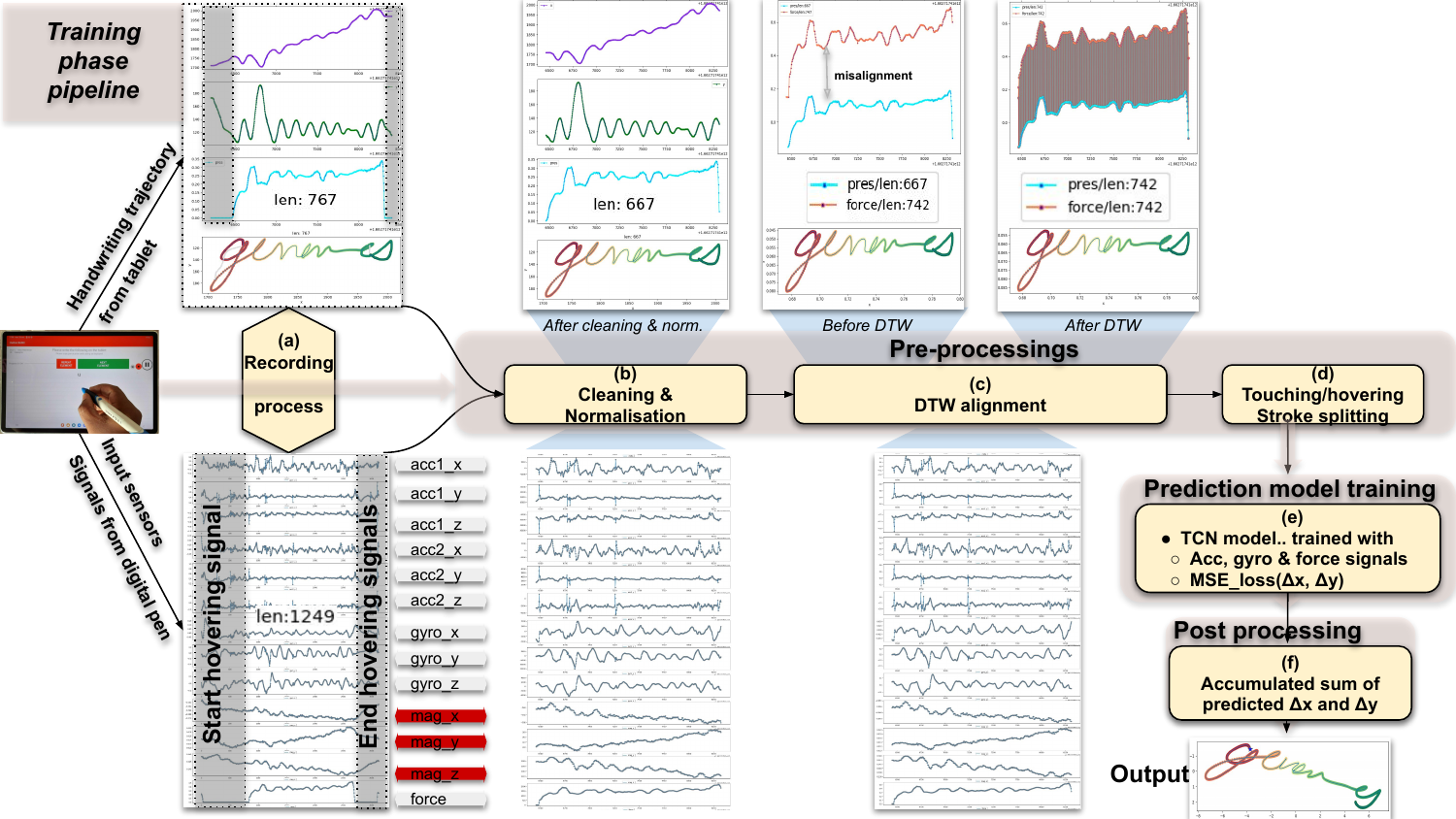}
\caption{Training processing steps of our proposed handwriting trajectory reconstruction pipeline.}
\label{fig:pipline-training}
\end{center}
\end{figure*}

\begin{figure*}[thpb]
\begin{center}
\includegraphics[scale=0.54]{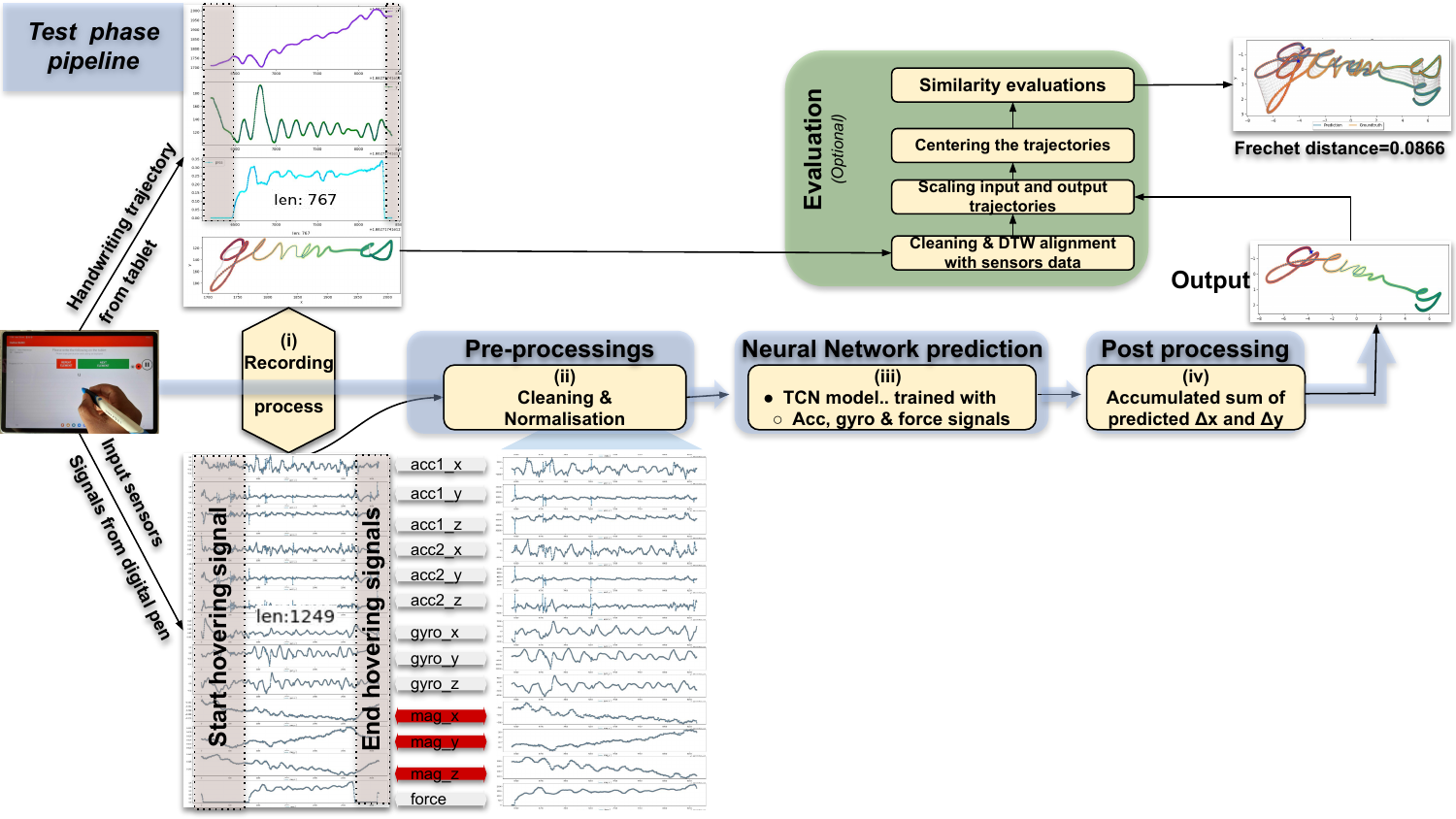}
\caption{Test processing steps together with post processing evaluation of the proposed handwriting trajectory reconstruction pipeline.}
\label{fig:pipline-testing}
\end{center}
\end{figure*}
To reconstruct the handwriting trajectory from IMU signals, we present a complete pipeline which consists of training and test phases as illustrated in Figures \ref{fig:pipline-training}, and \ref{fig:pipline-testing}.
The training phase consists in: (i) a preprossessing process that produces cleaned and aligned input sensors signals and output handwriting trajectory signals (Figure~\ref{fig:pipline-training}(b) to (d)); (ii) a reconstruction training process, where a neural network model based on a Temporal Convolutional Network (TCN) architecture is trained on strokes using a dedicated loss function to predict the displacement vectors of the handwriting trajectory (Figure~\ref{fig:pipline-training}(e)); 
(iii) a post-processing to obtain the handwriting trajectory signal from the expected displacement vectors. (Figure~\ref{fig:pipline-training}(f)). 

The test phase comprises the three principal steps of the training phase. However, preprocessing is reduced to only cleaning and normalizing the input signals (Figure~\ref{fig:pipline-testing}(ii)). In addition, the prediction model (Figure~\ref{fig:pipline-testing}(iii)) is applied on the whole input signal of the target sample, including touching and hovering strokes of the sample. \yann{Post-processing are the same as in the training phase followed by scaling and centering steps to compare the predicted trajectory with the ground truth.}

To the next, we describe the three main steps of our proposed pipeline: data processing, our neural network and post-processing. Every steps are discussed for training and test phases. 

\subsection{Data preprocessing}
Data prepossessing are required at the training phase due to the misalignment of the pen and tablet signals and to the noise in signals. 
First, signals are divided into spans that correspond to the written samples. One recalls that input signals captured from the Digipen are composed of 13 channels while signals from the tablet are made up of 3 channels (x, y and pressure). 
The timestamps of the Digipen and the tablet are added as an additional channel. 
Then, every sample is  passed forward to the following process section \ref{Cleaning and normalisation}.

\subsubsection{Cleaning and normalization}
\label{Cleaning and normalisation}
As part of the training phase, irrelevant parts of the input signals are removed. Those parts are the start and end pen-up movements that does not refer to pen-up and pen-down actions to write the given script (Figure~\ref{fig:pipline-training}(b)). 
Since the input sensor signals represent the relative values of the acceleration, speed and orientation of the Digipen, it is reasonable to map it to the displacement vectors of the handwriting trajectory signal rather than to the real values of handwriting trajectory.
Thus, displacement vectors ($\Delta$x, $\Delta$y) are computed from the (x, y) channels of the handwriting trajectory.
In order to maintain the interoperability of the system between the different versions of Digipen, accelerometers, gyroscope, magnetometer and force signals are normalized by their maximum values (as reported by the manufacturer). 
In the test phase, the cleaning and normalization process concerns the sensors' signal only (Figure~\ref{fig:pipline-testing}(ii)).

\subsubsection{Input - Output DTW alignment}
\yann{Due to the different sampling rates between the stylus and the tablet, an alignment process is crucial to get a mapping between the input and output time series for training.}  
In addition, the sensor data are acquired in packets of 6 data points and the recorded timestamps are not equally spaced due the Bluetooth transmission time delay. 
Due to this mismatch, the recorded input and output signals have different lengths and they are not synchronized.


\begin{figure*}[thpb]
\begin{center}
\includegraphics[scale=0.55]{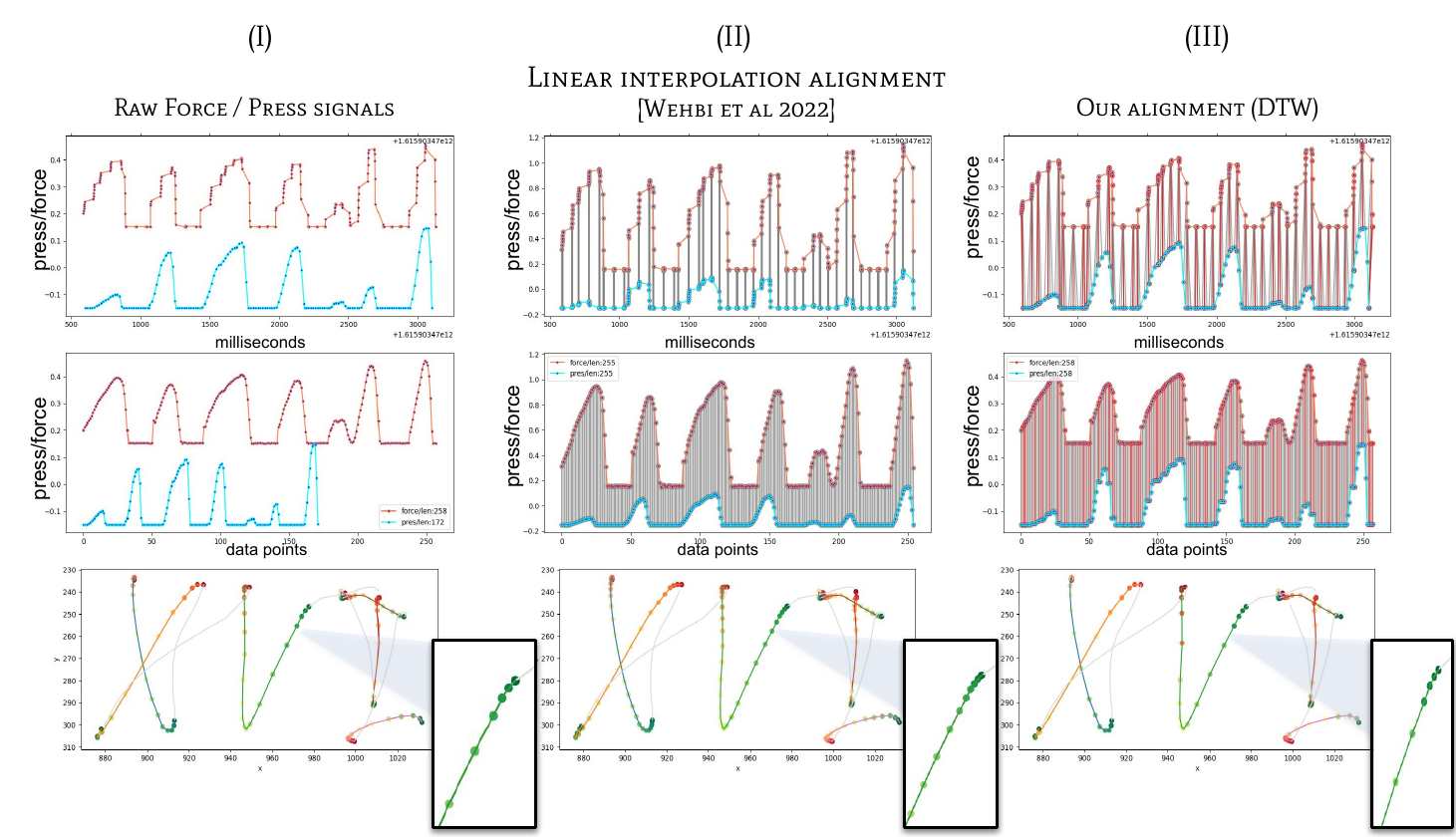} 
\caption{Our proposed DTW based alignment compared to the linear interpolation based alignment proposed by \cite{wehbi2022surface}. The left side image represents the raw force / pressure (sensor/online) signals, the middle image shows the interpolation alignment result and the right side image shows the DTW based alignment results. The red lines connecting the two signals represent the duplicate points added to the pressure signal and the grey ones represent the one-to-one alignment.}
\label{fig:align}
\end{center}
\end{figure*}

In order to respect the writing dynamics as much as possible, we propose an alignment approach based on the Dynamic Time Warping algorithm (DTW) (Figure~\ref{fig:pipline-training}(C)). 
The pen and tablet signals are recorded simultaneously. 
The transmission time delay is between 10 and 40 milliseconds in the Digipen version 6.0. 
Since the sampling rate of the sensor data is higher than the one of the tablet data, we need to up-sample the tablet data to match the sensors data length.

First, the average transmission time delay is subtracted to the tablet data in order to approximately synchronize it with sensors data. 
Then, we use the DTW algorithm (for more details see the appendix) to find an alignment path between the timestamps of the stylus and the tablet.

In practice, we observe that aligning the timestamps using DTW as we do is more effective than relying on the DTW alignment of force and pressure data.
Figure~\ref{fig:align} shows a comparison between the alignment approach used in \cite{wehbi2022surface} based on linear interpolation and our proposed approach based on DTW. Alignments are presented on the force (pen sensor) signal against the pressure signal (from the tablet). 
The icon images of Figure \ref{fig:align} shows that the DTW alignment (Figure~\ref{fig:align}(III)) results in the same data distribution than the raw points while the linear interpolation one (Figure~\ref{fig:align}(II)) spaces the data points equally.

\subsubsection{Splitting into strokes}
Splitting into strokes only concerns the training phase of the proposed pipeline.
It enables to train the neural network on the touching parts (strokes) only (Figure~\ref{fig:pipline-training}(d)). The idea is to use the force/pressure signals of the sensors and handwriting trajectory signals to determine the touching strokes of the training samples, knowing that we consider every touching stroke whose force/pressure values is greater than a predefined threshold, namely 0.01 for the force and 0 for the pressure. 

We have the choice to either learn the neural network model on the entire training sample including the touching and hovering stroke, or to learn it on the touching strokes only. We will investigate it in the experimental part. 
However, we always test the model on the entire test samples (including the touching and hovering strokes).

\subsection{Neural Network model}
\yann{We design a neural network architecture to predict the displacement vectors of the handwriting trajectory from the tablet, given a sensors signal of the Digipen (Figure~\ref{fig:pipline-training}(e)). The training and test steps of our proposed model are described in the following sub-sections.}

\subsubsection{Model architecture}
In order to take into account the past and future states when mapping an input sequence toward an output signal, we propose to use a non-causal Temporal Convolutional neural Network (TCN) architecture as in \cite{bai2018empirical}. The choice of a TCN architecture is supported by the great success of the TCN for sequence-to-sequence tasks with few training samples, e.g. for weather prediction~\cite{yan2020temporal}, traffic prediction~\cite{dai2020hybrid} and sound event localization and detection~\cite{guirguis2021seld}. 

The CNN architecture proposed in \cite{wehbi2022surface} captures the spatial features that refer to the arrangement of data points of a sequence, and the relationship between them within the sequence. However, it lacks the hierarchical and distant dependencies that can be grasped with dilated convolutions and a deep architecture \yann{as in a TCN}. 

It was proved that TCN architecture is most suitable to extract relevant spatial and temporal features for a sequence of frames describing an action compared to an LSTM recurrent network \cite{nan2021comparison}. \textcolor{blue}{\romain{Indeed, recurrent architectures are known to suffer from vanishing gradient problems in very long sequences, such as those produced by IMU sensors. Thus, TCN based systems outperforms their LSTM counterparts in different fields of application such as anomaly detection \cite{gopali2021comparative} or skeleton-based action recognition \cite{nan2021comparison}.}}

\begin{figure}[!t]
\begin{center}
\includegraphics[scale=0.4]{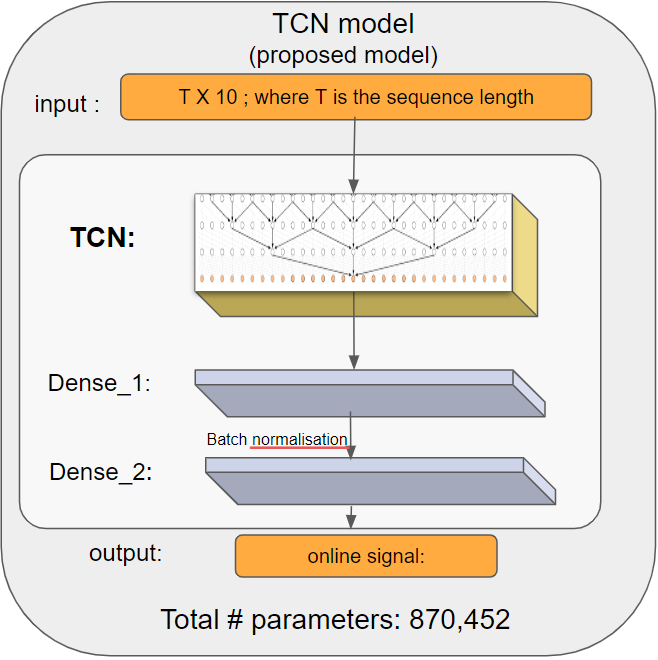}
\caption{Our TCN model for handwriting trajectory reconstruction.}
\label{fig:TCN}
\end{center}
\end{figure}

Our model receives 10 of 13 channels of the sensor data. Magnetometer data are neglected due to the interference of the tablet magnetic field with the signal.
Figure~\ref{fig:TCN} illustrates our network architecture. 
It is composed of TCN-like layers, a dense layer followed by a batch-normalization layer before the last dense layer that produces the output of the network (the two ($\Delta$x, $\Delta$y) channels).
The TCN part consists of three stacked inner blocks of non-causal and dilated 1D-convolutions with kernel size of 5. The dilation rate applied to each block is increased with the depth of the network, from 1 to 4 
and every inner layer is followed by a batch normalization layer. \yann{More details are given in appendix. }


\subsubsection{Model training and test}

During the training phase, the model is trained to minimize the Mean Squared Error between the real and predicted displacement vectors of the handwriting trajectories. One uses the ADAM optimization, a batch size of 16 and a learning rate of $10e^{-3}$. 
As stated before, the 10 input channels of the sensor signal are the front and rear accelerators, gyroscope and force channels of the sensor signal, obtained after the cleaning and normalization process.

During the test phase, a cleaned and normalized signal is given as input of the model that predict the corresponding displacement vectors of the handwriting trajectory signal. 

\subsection{Post-processing}
This process is the same in training and test phases. 
It consists in reconstructing the predicted handwriting trajectories by computing the accumulated values from the predicted displacement vectors (Figure~\ref{fig:pipline-training}(f), and Figure~\ref{fig:pipline-testing}(iv)). 

\yann{At test time, to evaluate the shape of the prediction, we normalize and center the predicted trajectory and the reference trajectory using the Procrustes analysis to fit a uniform scale. }



\section{Experiments and results}
This section presents the experimental framework, especially the evaluation protocol and the metric used to evaluate our model. Then, we investigate the benefit of the successive parts of our approach. First, we experiment variants of networks to explore various receptive field sizes. Second, one compares our pipeline to a previous approach \cite{wehbi2022surface} and evaluates the benefit of using both our alignment method and our model. Finally, one explores the impact of training our model both on touching and hovering strokes compared to training it on touching strokes only, when the goal is always to predict both touching and hovering strokes. 

\subsection{Evaluation protocol and metric}

\yann{Some works perform trajectory reconstruction as a step towards recognition. Thus, final recognition performance can be used as an evaluation criterion~\cite{wehbi2022surface} and, as discussed before, it cannot reflect well the quality of the reconstruction. Here, the end goal is the trajectory reconstruction and dedicated metrics should be used to assess similarity between ground truth and reconstructed trajectories.}
\textcolor{blue}{ 
\romain{
\wassim{\Florent{Root Mean Squared Error (RMSE) is a natural candidate~\cite{ott2022joint}, but it can't handle temporal alignment. Dynamic Time Warping (DTW), allow this temporal alignment.} 
Chen et al.~\cite{chen2022complex} created the Adaptive Intersection over Union (AIOU) metric, which uses the stroke width and thickness as well as the length-independent DTW to compare the trajectory-point alignment of two different handwriting trajectories of different lengths.}
In this work, we decide to rely on the Fréchet distance , \florent{whose formal definition is given in appendix section \ref{sec:Frechet}}. This distance has benefits that we detail in the appendix section \ref{sec:Frechet}.  \yann{The Fréchet distance measures the distance between curves, by taking into account the location and ordering of the points along the curves \cite{har2002new}.} This allows to capture both local and global information accurately, and correlates well with qualitative assessment in practice.
}}



\florent{In fact, due to the limited tracking capacity of the tablet over 15mm above the tablet screen, the handwriting signal is partially captured on hovering. For that, \yann{we only consider the touching parts (pen-down) for which we can compare a trajectory prediction to the handwriting trajectory from the tablet and compute the Fréchet distance.}} 

\wassim{We used the FAU-EINNS dataset, our IRISA-KIHT dataset, and a subset of it, IRISA-KIHT-S, to compare our processing pipeline to \cite{wehbi2022surface}}
\wassim{IRISA-KIHT and IRISA-KIHT-S contain 38 and 30 writers' recordings, respectively. Each recording contains 34 characters, words, word groups, equations, and shapes. Section \ref{datasets description} of the appendix describes the datasets.}
The FAU-EINNS dataset consists of words written by 6 writers, and we take the samples of the user numbers 1, 2, 3, 5 and 6 for training (1774 samples) and testing on the 344 samples. 

\subsection{Receptive field effect}

\begin{table*}
\centering
\footnotesize
\caption{Fréchet distance with standard deviation from TCN model with varied receptive fields on FAU-EINNS dataset}
\label{vrf_Mbase}
\begin{tabular}{lcccc}
Type   & \begin{tabular}[c]{@{}c@{}}TCN-49\end{tabular} & \begin{tabular}[c]{@{}c@{}}TCN-85\end{tabular} & \begin{tabular}[c]{@{}c@{}}TCN-169\end{tabular} & \begin{tabular}[c]{@{}c@{}}TCN-373\end{tabular}  \\ 
\hline
Global & \textbf{0.077} $\pm $ 0.0356                                             & 0.0867 $\pm $ 0.0388                                                    & 0.0893 $\pm $ 0.0426                                                      & 0.0966 $\pm $ 0.0335                                                       \\
\hline
\end{tabular}
\end{table*}

\begin{table*}[ht]
\centering
\footnotesize
\caption{Fréchet distance with standard deviation from TCN model with varied receptive fields on IRISA-KIHT dataset}
\label{vrf_ourbase}
\begin{tabular}{lcccc}
Type       & \begin{tabular}[c]{@{}c@{}}TCN-49\end{tabular} & \begin{tabular}[c]{@{}c@{}}TCN-85\end{tabular} & \begin{tabular}[c]{@{}c@{}}TCN-169\end{tabular} & \begin{tabular}[c]{@{}c@{}}TCN-373\end{tabular}  \\ 
\hline
Global     
& \textcolor{black}{0.2039 $\pm $  0.1427}                                                     
& \textcolor{black}{0.2415 $\pm $  0.1676}                                                     
& \textcolor{black}{0.2042 $\pm $  0.1419}                                                      
& \textcolor{black}{\textbf{0.1807} $\pm $  0.1338}                                              \\ 
\hline
Characters 
& \textcolor{black}{0.2758 $\pm $ 0.1709}                                                     
& \textcolor{black}{0.3048 $\pm $ 0.2087}                                                     
& \textcolor{black}{\textbf{0.236} $\pm $ 0.1901}                                            
& \textcolor{black}{0.2402 $\pm $ 0.1642}                                                       \\ 
\hline
Words      
& \textcolor{black}{0.1383 $\pm $ 0.0727}                                             
& \textcolor{black}{0.1716 $\pm $ 0.0886}                                                      
& \textcolor{black}{0.1605 $\pm $ 0.0627}                                                      
& \textcolor{black}{\textbf{0.1207} $\pm $ 0.0702}                                                       \\ 
\hline
Equation   
& \textcolor{black}{\textbf{0.1438} $\pm $ 0.0568}                                            
& \textcolor{black}{0.213 $\pm $ 0.0998}                                                     
& \textcolor{black}{0.1964 $\pm $ 0.0639}                                                      
& \textcolor{black}{0.1462 $\pm $ 0.0584}                                                       \\ 
\hline
Shapes     
& \textcolor{black}{0.281 $\pm $ 0.1071}                                                     
& \textcolor{black}{0.3069 $\pm $ 0.1289}                                                     
& \textcolor{black}{0.2758 $\pm $ 0.12}                                              
& \textcolor{black}{\textbf{0.2505} $\pm $ 0.1195}                                                       \\ 
\hline
Word groups  
& \textcolor{black}{0.1552 $\pm $ 0.0665}                                                     
& \textcolor{black}{0.1971 $\pm $ 0.0775}                                                     
& \textcolor{black}{0.2028 $\pm $ 0.0843}                                                      
& \textcolor{black}{\textbf{0.1269} $\pm $ 0.0522}                                              \\
\hline
\end{tabular}
\end{table*}

To design the TCN model architecture, we pay attention to the sampling rate of the input signal of sensors. 
It is obvious that the TCN based neural network may have the capacity to absorb the noise of such low quality and noisy signals, depending on the size of its receptive field. 

In order to evaluate the receptive field effect of our model, we trained and evaluated the model with different sizes of receptive fields equal to 49, 85, 168 and 373 on both FAU-EINNS (100Hz) and IRISA-KIHT (400Hz) datasets.

Table \ref{vrf_Mbase} shows that the TCN model with the smallest receptive field size (49) performs better than the ones with greater receptive fields on FAU-EINNS (100Hz) dataset. 
On our IRISA-KIHT (400Hz) dataset, the TCN model with the greatest receptive field (373) generally outperforms the other sizes of receptive field as seen in Table \ref{vrf_ourbase}. 
With a big receptive field, a large context can be exploited to make the prediction as well as the signal noise of long hovering can be absorbed, with the cost of higher number of learnable parameters.
\begin{table}[!b]
\centering
\footnotesize
\caption{TCN models' size with various receptive fields}
\begin{tabular}{lllll}
\# Params & TCN-49 & TCN-85 & TCN-169 & TCN-373  \\ 
\hline
Total      & 467,452 & 528,452 & 870,452 & 894,452 \\
Trainable   & 464,152 & 524,752& 866,752 & 888,352  \\
\hline
\label{param}
\end{tabular}
\end{table}
Table \ref{param} illustrates the size of TCN models in terms of number of learnable parameters with regard to the receptive fields.

\begin{figure*}[ht]
\begin{center}
\includegraphics[scale=0.3]{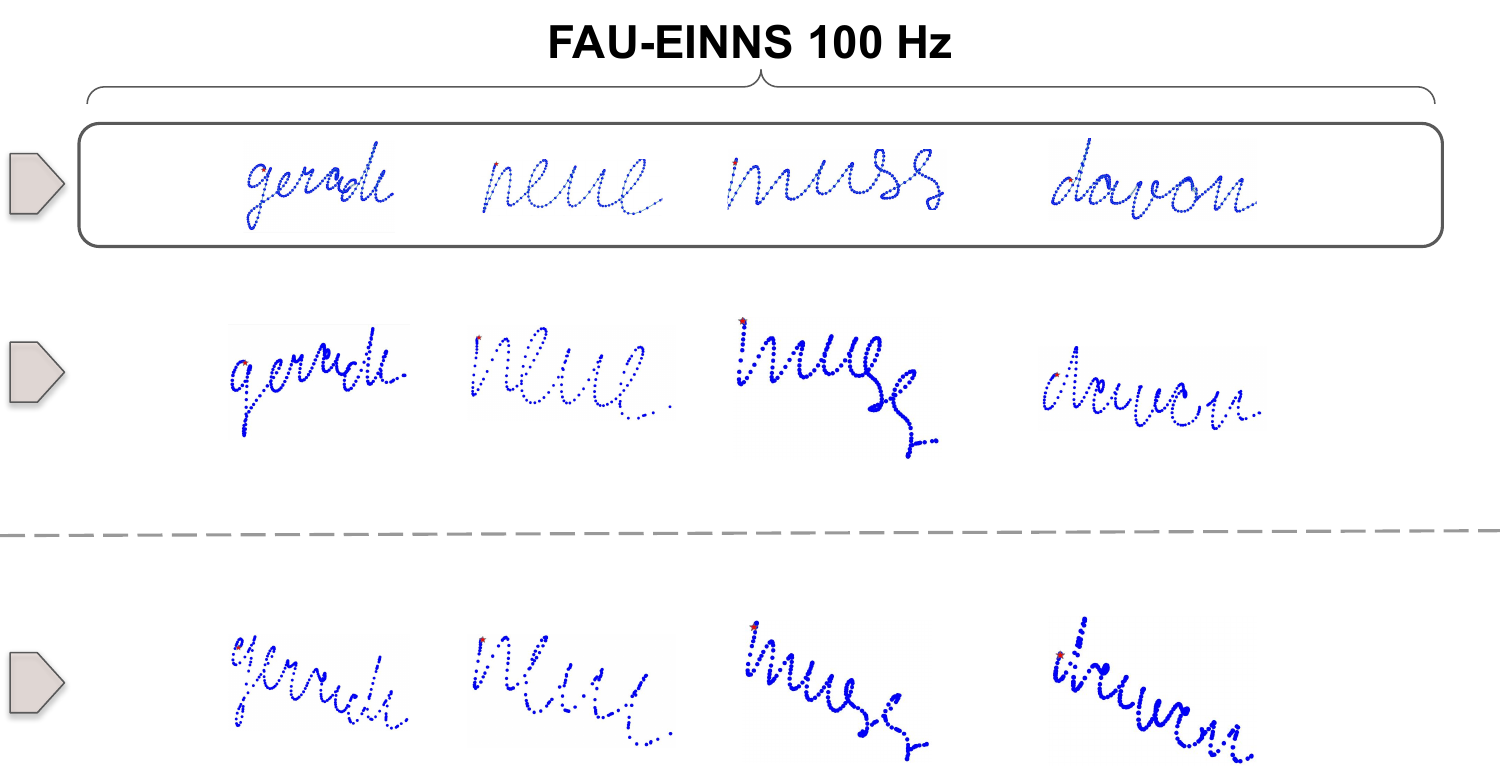}
\includegraphics[scale=0.31]{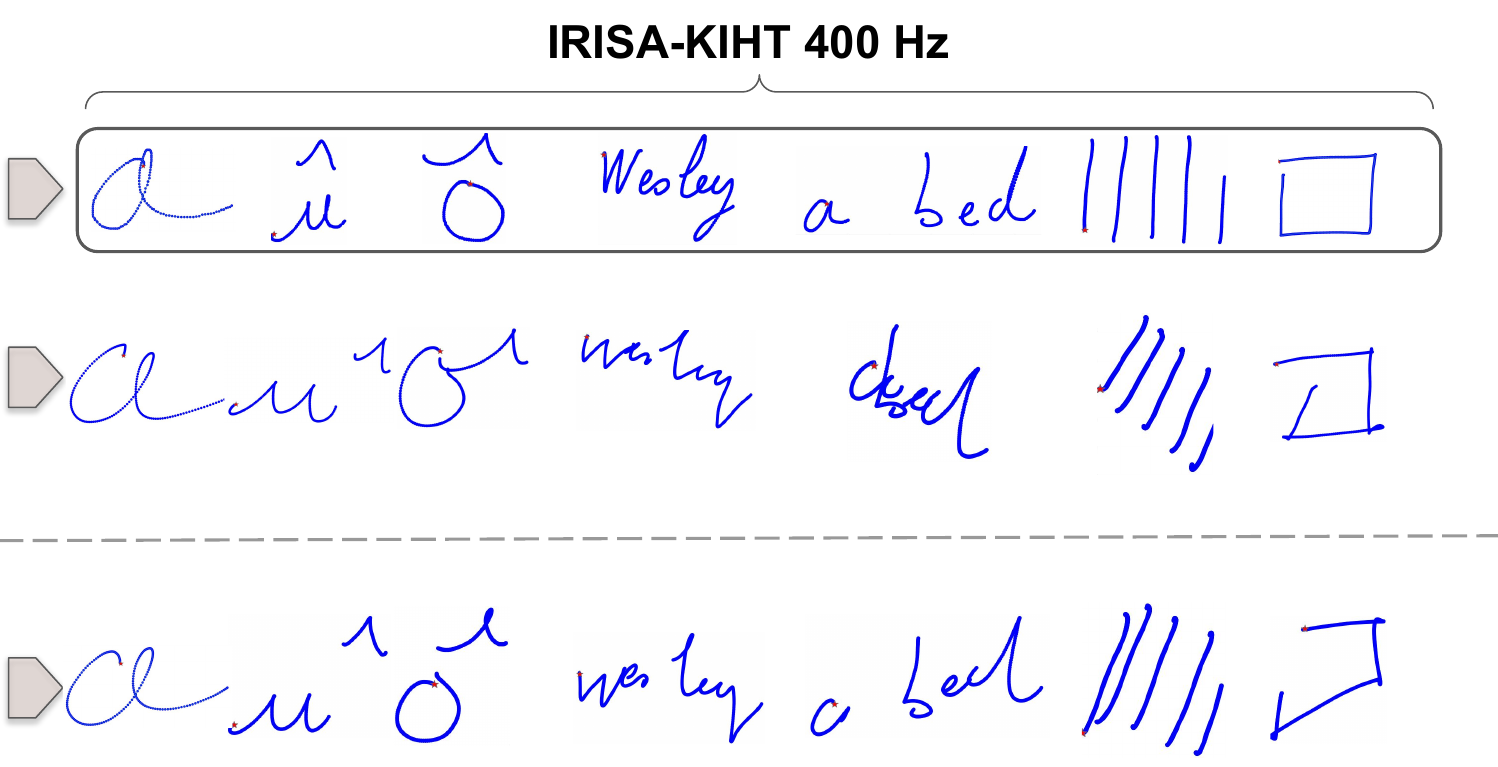}
\caption{Handwriting reconstructions with two TCN models with different receptive fields on the FAU-EINNS (left) and IRISA-KIHT (right) datasets. First line, ground-truth. Second, the 49-receptive-field TCN model. Third line: TCN model with 373 receptive field.}
\label{fig:rf}
\end{center}
\end{figure*}

By comparing TCN-49 and TCN-373, during hovering strokes, the network may see the last and next touching points of two successive touching strokes, as illustrated in Table \ref{vrf_ourbase} with better results on word groups. These two models are quite similar in terms of performance (Figure \ref{fig:rf}). As one of our next goal is to embed the model inside the Digipen, we decided to keep the TCN-49 model to get a good trade-off between the model performance and the number of parameters. 

\begin{table}[ht]
\centering \footnotesize
\caption{Average Fréchet distance with standard deviation on IRISA-KIHT-S over 3-fold compared to \cite{wehbi2022surface}}
\label{OurVsSOA_eval_fold}
\begin{tabular}{lcc} 
\hline
           & \begin{tabular}[c]{@{}c@{}} CNN model \cite{wehbi2022surface}\end{tabular} & \begin{tabular}[c]{@{}c@{}}Our approach\end{tabular}  \\ 
\hline 
\textcolor{black}{Fold 1} 
& \textcolor{black}{0.4055 $\pm $ 0.2699  }                                                                      & \textcolor{black}{\textbf{ 0.2099} $\pm $ 0.1750}                                                         \\ 
\hline 
\textcolor{black}{Fold 2  }
& \textcolor{black}{\multirow{1}{*}{0.5188 $\pm $ 0.3926}  }                                                     & \textcolor{black}{\multirow{1}{*}{\textbf{0.3122} $\pm $ 0.1731} }                                        \\ 
\hline  
\textcolor{black}{Fold 3  }
& \textcolor{black}{\multirow{1}{*}{0.4383 $\pm $ 0.3701} }                                                      & \textcolor{black}{\multirow{1}{*}{\textbf{0.1766} $\pm $ 0.1056}}   \\ 
\hline  
\textcolor{black}{Mean}  
& \textcolor{black}{\multirow{1}{*}{0.4542 $\pm $ 0.3442} }                                                      & \textcolor{black}{\multirow{1}{*}{\textbf{0.2329} $\pm $ 0.1512}  }

\\ 
\hline
\end{tabular}
\end{table}

\begin{table}[ht]
\centering \footnotesize
\caption{Average Fréchet distance with standard deviation of our pipeline compared to \cite{wehbi2022surface}}
\label{OurVsSOA_eval}
\begin{tabular}{lcc} 
\hline
           & \begin{tabular}[c]{@{}c@{}} CNN model \cite{wehbi2022surface}\end{tabular} & \begin{tabular}[c]{@{}c@{}}Our approach\end{tabular}  \\ 
\hline 
FAU-EINNS & 0.2628 $\pm $ 0.1256                                                                        & \textbf{0.077} $\pm $ 0.0356                                                           \\ 
\hline 
IRISA-KIHT 
& \textcolor{black}{\multirow{1}{*}{0.4691 $\pm $ 0.2465} }                                                       & \textcolor{black}{\multirow{1}{*}{\textbf{0.2039} $\pm $ 0.1427}  }                                        \\ 
\hline  
IRISA-KIHT-S 
& \textcolor{black}{\multirow{1}{*}{0.4542 $\pm $ 0.3442} }                                                       & \textcolor{black}{\multirow{1}{*}{\textbf{0.2329} $\pm $ 0.1512} }                                          \\ 
\hline
\end{tabular}
\end{table}

\subsection{Comparison with related work}
\romain{We compare the performance of our approach against the one of Wehbi et al. \cite{wehbi2022surface} on both FAU-EINNS dataset and our own dataset, called IRISA-KIHT. We also provide results on the subset of the latter called IRISA-KIHT-S which is made publicly available and can hence be used as benchmark for future works. We applied the same evaluation protocol on both models and processing pipelines. 
Due to the comparatively smaller size of IRISA-KIHT-S, a 3 fold cross validation is used when this dataset is at stake, as shown in Table \ref{OurVsSOA_eval_fold}.}
Results show that our pipeline (TCN-49 model and DTW based alignment) outperforms Wehbi et al.\cite{wehbi2022surface} approach (CNN model and linear interpolation alignment) on every dataset, both quantitatively as seen in Table~\ref{OurVsSOA_eval} on the average Fréchet distance and qualitatively as illustrated in Figure~\ref{fig:approcheSeval} (Wehbi et al.  \cite{wehbi2022surface} appears on the second line of the figure while our approach appears on the bottom line). 
\romain{Despite the difference in size between IRISA-KIHT and IRISA-KIHT-S datasets, one can observe that reported performance is comparable in both datasets for both our method and its competitor (see table~\ref{OurVsSOA_eval}).
This confirms that IRISA-KIHT-S is of sufficient size for the quantitative assessment of OH trajectory reconstruction from IMU sensor data.}
In the next section, we will evaluate each step of the processing chain. 

\subsection{Alignment methods and models}
In the following, we compare two effective processes of the handwriting trajectory reconstruction pipeline; (i) the input-output alignment process and (ii) the reconstruction model performance.
We provide a cross validation scheme between the two alignment methods (linear interpolation and DTW based alignment methods) and the two reconstruction models (CNN and TCN based models) on the FAU-EINNS and IRISA-KIHT datasets.
From Table \ref{OurVsM_eval}, we observe that each step of our approach outperforms the counterpart of the Wehbi et al. \cite{wehbi2022surface} method on both datasets. 
The results show that our alignment is better whatever the model on all the datasets.

\begin{figure*}[thpb]
\begin{center}
\includegraphics[scale=0.3]{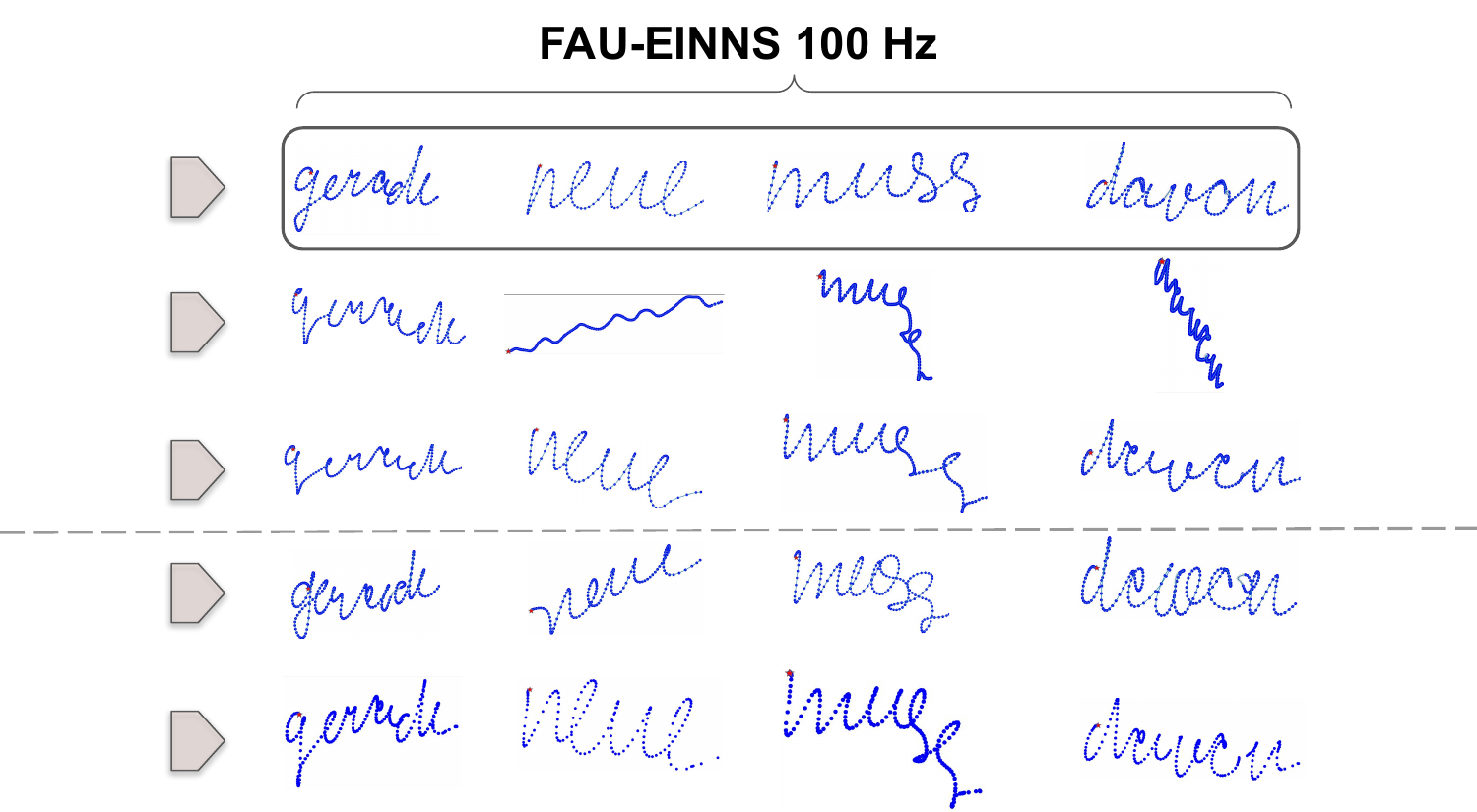}
\includegraphics[scale=0.29]{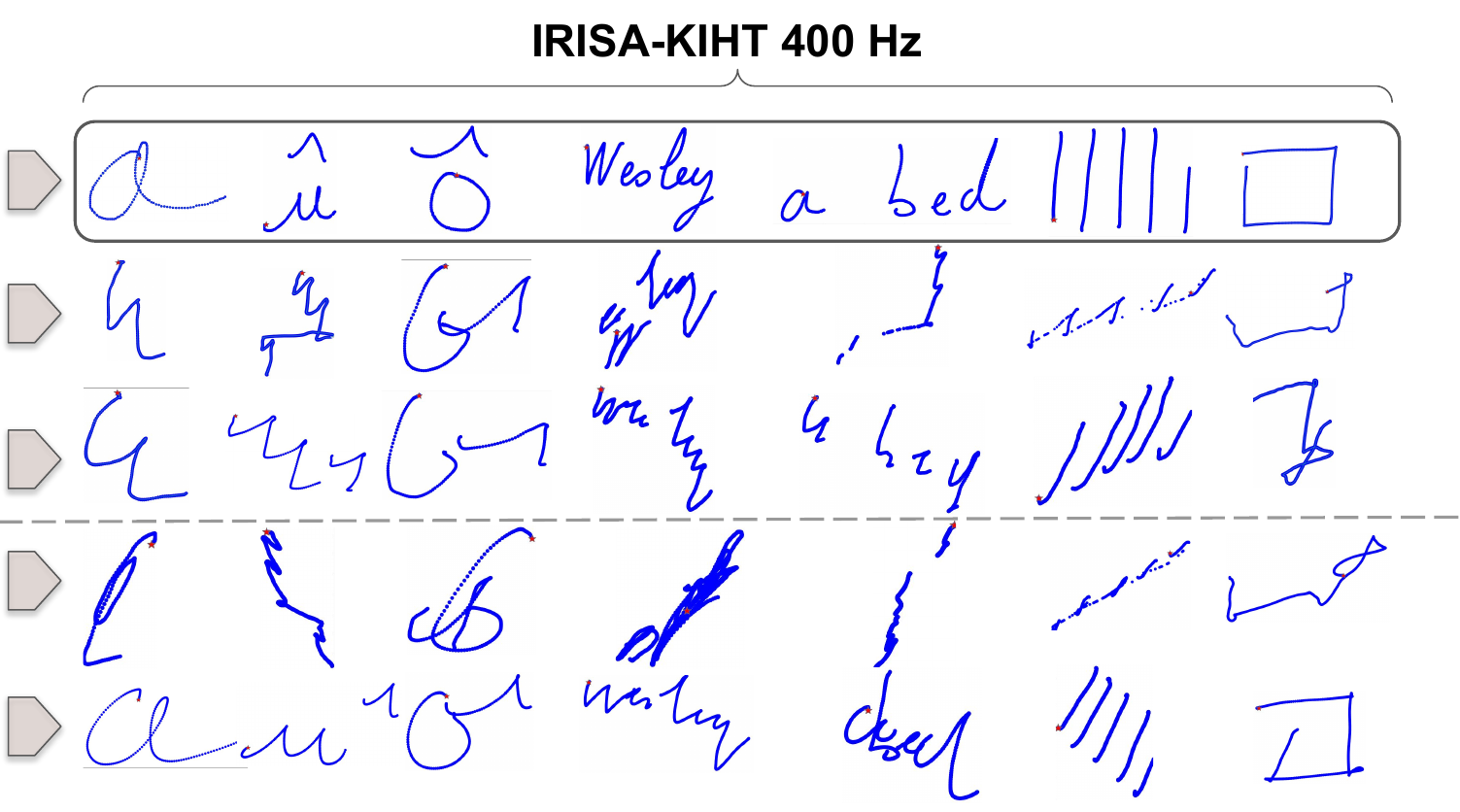}
\caption{Comparison with \cite{wehbi2022surface} approach with our one. First line, \cite{wehbi2022surface} CNN model and linear interpolation alignment results. Second line, CNN model and the introduced DTW alignment results. Third line, our TCN model and linear interpolation alignment results. Last line, our TCN model and the proposed DTW alignment results.
   }
\label{fig:approcheSeval}
\end{center}
\end{figure*}

In contrast to the linear alignment, our alignment based on the DTW algorithm keeps the writing dynamic, which seems to be essential to reach quality  trajectory reconstruction.   
Figure~\ref{fig:approcheSeval} shows examples of reconstruction trajectories using the CNN or TCN models, when using the linear interpolation and the DTW alignments on the two datasets. Moreover, our TCN model outperforms the CNN counterpart whatever the alignment method, which shows that the TCN model benefits from a larger receptive field and a deeper network to model more complex patterns and to be less sensitive to the noise. 

\begin{table*}
\centering \footnotesize
\caption{Model and alignment method comparison between \cite{wehbi2022surface} and ours on both datasets.}
\begin{tabular}{lllll} 
\cline{2-5}                                                                            &              \multicolumn{2}{c}{CNN model \cite{wehbi2022surface}}                                                                                                        & \multicolumn{2}{c}{Our TCN model-49}                                                                                                         \\ 
\cline{2-5}
                                                                          &               \begin{tabular}[c]{@{}l@{}}Linear \\interpolation \\alignment \cite{wehbi2022surface}\end{tabular} & \begin{tabular}[c]{@{}l@{}}Our DTW \\alignment\end{tabular} & \begin{tabular}[c]{@{}l@{}}Linear \\interpolation \\alignment \cite{wehbi2022surface}\end{tabular} & \begin{tabular}[c]{@{}l@{}}Our DTW \\alignment\end{tabular}  \\ 
\hline
 FAU-EINNS 100Hz
      & 0.2628 $\pm $ 0.1256 
                    & 0.0801 $\pm $ 0.0352  
                    & 0.1211 $\pm $ 0.0639 
                    & \textbf{0.077} $\pm $ 0.0356 \\ 
\hline 
IRISA-KIHT 400Hz
    & \textcolor{black}{0.4691 $\pm $ 0.1427} 
                    & \textcolor{black}{0.2095 $\pm $  0.1255} 
                    & \textcolor{black}{0.3859 $\pm $ 0.2003} 
                    & \textcolor{black}{\textbf{0.2039} $\pm $ 0.1427}  \\
\hline
\label{OurVsM_eval}
\end{tabular}
\end{table*}

\subsection{Touching versus hovering trajectories reconstruction}
\begin{figure}[!b]
\begin{center}
\includegraphics[scale=0.5]{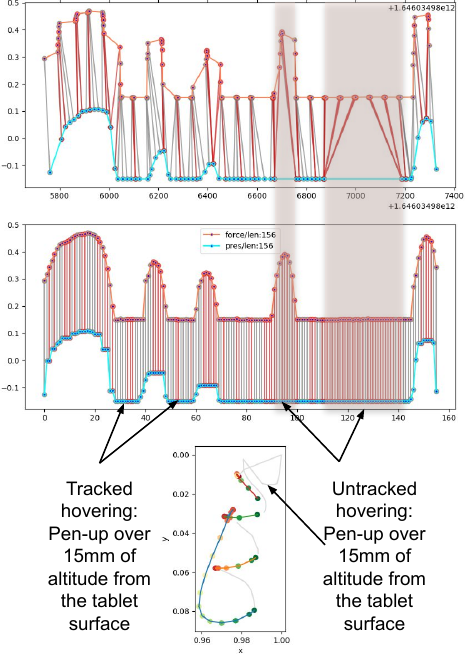}
\caption{Tracked hovering and untracked hovering effects on the alignment process}
\label{fig:hoveringeffect}
\end{center}
\end{figure}

\begin{figure}[th]
\begin{center}
\includegraphics[scale=0.29]{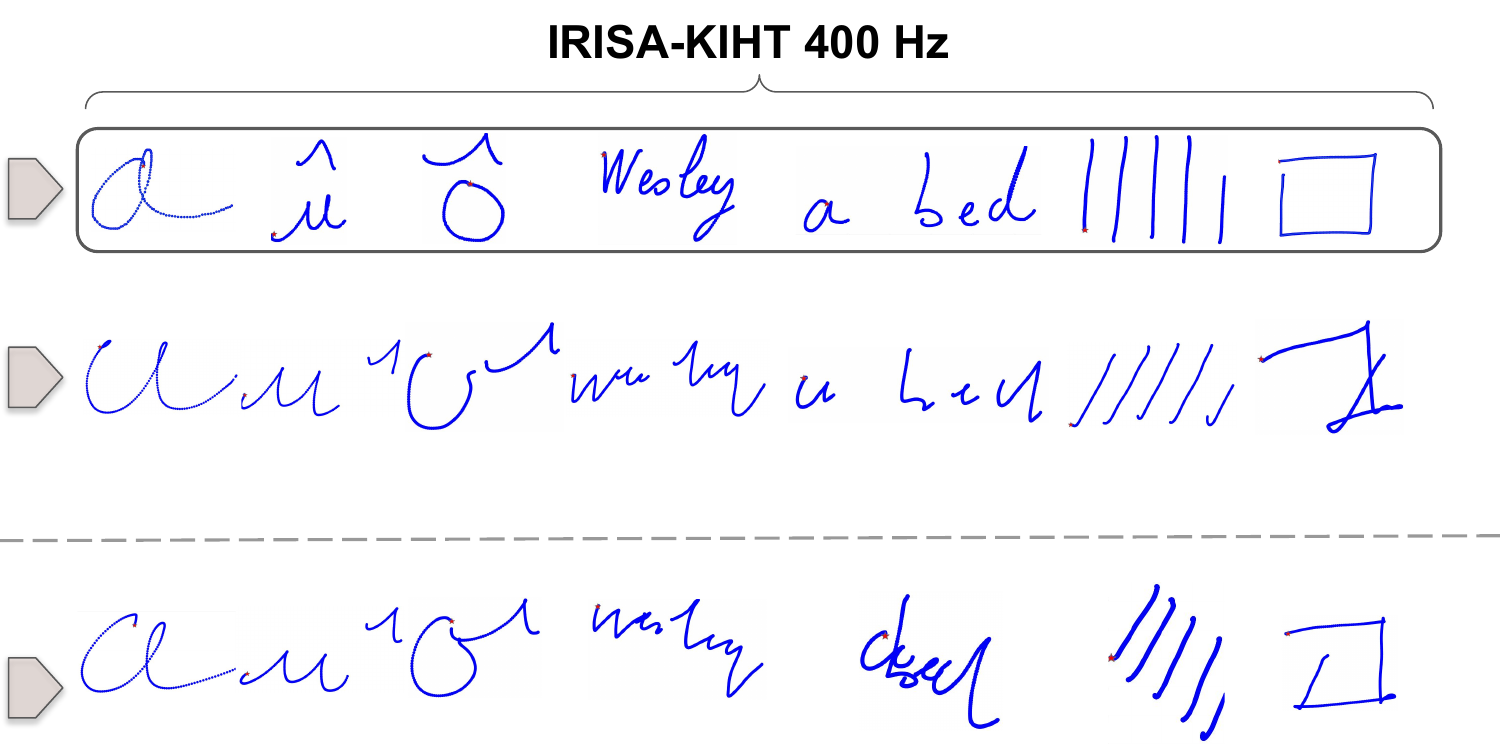}
\end{center}
\caption{TCN-49 model performance when training on hovering and touching data (middle) and touching data only (bottom). The ground truth is on first line.}
\label{fig:touchinghovering}
\end{figure}

\begin{table}[th]
\centering
\footnotesize
\caption{Training on touching strokes versus learning on touching \& hovering strokes results }
\label{tbl:tch+hvr}
\begin{tabular}{lcc}
         & \multicolumn{2}{c}{Our TCN Model-49}                                                                                                                                   \\ 
\cline{2-3}
Type  & \begin{tabular}[c]{@{}c@{}}Trained on touching \\\& hovering strokes\end{tabular} & \begin{tabular}[c]{@{}c@{}}Trained on touching\\~strokes only\end{tabular}  \\ 
\hline
Global 
& \textcolor{black}{0.2471 $\pm $ 0.1711}   
& \textcolor{black}{\textbf{0.2039} $\pm $ 0.1427}      \\
\hline
Characters 
& \textcolor{black}{0.3587 $\pm $ 0.185}   
& \textcolor{black}{\textbf{0.2758} $\pm $ 0.1709}     \\
\hline
Words 
& \textcolor{black}{0.1497 $\pm $ 0.0671}   
& \textcolor{black}{\textbf{0.1383} $\pm $ 0.0727}     \\
\hline
Equations 
& \textcolor{black}{0.1483 $\pm $ 0.0666}   
& \textcolor{black}{\textbf{0.1438} $\pm $ 0.0568} \\
\hline
Shapes 
& \textcolor{black}{0.429 $\pm $ 0.2094}   
& \textcolor{black}{\textbf{0.281} $\pm $  0.1071}     \\
\hline
Word groups 
& \textcolor{black}{\textbf{0.1497} $\pm $ 0.0696}   
& \textcolor{black}{0.1552  $\pm $ 0.0665}     \\
\hline
\end{tabular}
\end{table}

Recalling the definition of the touching and hovering strokes, a touching stroke consists of data points that are recorded while touching the tablet screen. Hence, hovering stroke is recorded partially due to the limited capacity of the tablet screen to track the pen when it goes up to 15mm\footnote{http://tennojim.xyz/article/wacom\_intuos\_pro\_l\_guide}.
As a result, only the Digipen continue to provide sensor signals when the pen is up over 15mm. 
In conclusion, there is no ground truth for such hovering movements. To offset the effect of over hovering on the data preprocessing, the proposed DTW alignment duplicates the end points of the period when the pen is hovering. Figure~\ref{fig:hoveringeffect} illustrates the effect of over hovering on the alignment process.

As illustrated in Figure~\ref{fig:hoveringeffect}, we distinguish two kinds of hoverings; (i) tracked hoverings, where the Digipen is still within the tracking field of the tablet. This kind of hovering is represented by the flat doted lines of Figure~\ref{fig:hoveringeffect}. (ii) untracked hoverings, when the Digipen goes too far from the tracking field of the tablet and consequently leading to lose the OH signal to be reconstructed by the model. 
At the same time, we observe a dual effect of missing the handwriting signal that can be correlated with the writer experience in writing. In fact, a fluent writer (adult) don't need to search where to start writing the next stroke of the same character or words. As a result, he don't need to raise the pen so high (outside the tablet' tracking field).
However, for sentences, the user needs to space words and he may raise the pen higher while searching where to put the pen on the screen again.

To avoid the untracked hoverings in the training datasets, we used the touching data only to train our model. The idea is to train the model on touching strokes only and test on touching and hovering strokes as well. We compared our model when it is trained on touching only or on touching and hovering strokes.  

Table~\ref{tbl:tch+hvr} shows that the strategy of learning on touching strokes only is generally better than the one of learning on touching and hovering strokes upon the distance of Frechet, and this is also the case for characters, words, equations and shapes categories. 
Results on the sentence category are very close. We think this is due to the correlated effect of the untracked hovering and the pen-movement hesitations of the writer between the words of the sentence.  
However, by looking at Figure~\ref{fig:touchinghovering}, we observe that the touching model reconstructs better touching strokes that appear in the form of single character (e.g. "a") or shape (e.g. "rectangle"). It is also better with double strokes characters (e.g. "û", "ô") where a single tracked hovering can be observed.
Similarly, we observe that our model reconstructs quite good the touching parts of the word groups but it fails to find where to start the reconstruction of the next stroke. This may happen due to untracked hoverings (that represents movement hesitations about where to put the pen again on the screen) like in "a --- bed" where there is a long untracked hovering between "a" and "bed".


\section{Conclusion \& perspectives}
This work introduces a novel processing pipeline for reconstructing handwriting trajectories from kinematic sensor signals. Input signals come from a digital pen designed by the Stabilo company called Digipen.
Our approach consists in training and testing phases, each one composed of three main processes; preprocessing, a neural network training/prediction process and  post-processing. A new dataset has been collected using a mobile application, a tablet and a Digipen and we provide benchmark results on it. 
To tackle with the sampling rate differences between the Digipen and the tablet, as well as the synchronization problem, we introduced a DTW based alignment approach. 
We show through experiments on several datasets that the TCN neural network architecture outperforms the CNN one proposed in Wehbi et al. \cite{wehbi2022surface} as well as the DTW based alignment approach that enhances the performance of both CNN and TCN models.
We also show that the model predicts better the touching and tracked hovering parts, when the model is trained on the touching strokes only.

\Florent{The major limitation of our approach is the modelling of the untracked / complex hovering trajectories that is still an open research problem. This constitutes a first important track for future works and could be solved using a semi-supervised or unsupervised deep learning framework.}
Moreover, in this paper, we only evaluated the shape of the reconstruction. We could extend our evaluation to the dynamics of the trajectory reconstruction such as speed, acceleration, etc. 



\section*{Acknowledgments}
This project is financed by the KIHT French-German bilateral ANR-21-FAI2-0007-01 project and these four partners, IRISA, KIT, Learn \& Go and Stabilo. This work was performed using HPC resources from GENCI-IDRIS (Grant 2021-AD011013148)

\bibliographystyle{plain}
\bibliography{sn-bibliography}

@article{pan2018handwriting,
  title={Handwriting trajectory reconstruction using low-cost imu},
  author={Pan, T-Y. and Kuo, C-H. and Liu, H-T. and others},
  journal={IEEE Transactions on Emerging Topics in Computational Intelligence},
  volume={3},
  number={3},
  pages={261--270},
  year={2018},
  publisher={IEEE}
}

@INPROCEEDINGS{7574685,
  author={Pan, T-Y. and Kuo, C-H. and Hu, M-C.},
  booktitle={2016 IEEE International Conference on Multimedia \& Expo Workshops (ICMEW)}, 
  title={A noise reduction method for IMU and its application on handwriting trajectory reconstruction}, 
  year={2016},
  volume={},
  number={},
  pages={1-6},
  doi={10.1109/ICMEW.2016.7574685}}

@inproceedings{ott2022joint,
  title={Joint Classification and Trajectory Regression of Online Handwriting using a Multi-Task Learning Approach},
  author={Ott, F. and R{\"u}gamer, D. and Heublein, L. and others},
  booktitle={Proceedings of the IEEE/CVF Winter Conference on Applications of Computer Vision},
  pages={266--276},
  year={2022}
}

@article{ott2022benchmarking,
  title={Benchmarking Online Sequence-to-Sequence and Character-based Handwriting Recognition from IMU-Enhanced Pens},
  author={Ott, F. and R{\"u}gamer, D. and Heublein, L. and others },
  journal={arXiv preprint arXiv:2202.07036},
  year={2022}
}

@inproceedings{wehbi2021towards,
  title={Towards an IMU-based Pen Online Handwriting Recognizer},
  author={Wehbi, M. and Hamann, T. and Barth, J. and others},
  booktitle={International Conference on Document Analysis and Recognition},
  pages={289--303},
  year={2021},
  organization={Springer}
}

@article{mustafid2022iamonsense,
  title={IAMonSense: Multi-level Handwriting Classification using Spatio-temporal Information},
  author={Mustafid, A. and Younas, J. and Lukowicz, P. and others},
  year={2022}
}

@article{ott2020onhw,
  title={The onhw dataset: Online handwriting recognition from imu-enhanced ballpoint pens with machine learning},
  author={Ott, F. and Wehbi, M. and Hamann, T. and others},
  journal={Proceedings of the ACM on Interactive, Mobile, Wearable and Ubiquitous Technologies},
  volume={4},
  number={3},
  pages={1--20},
  year={2020},
  publisher={ACM New York, NY, USA}
}

@article{wegmeth2021detecting,
  title={Detecting Handwritten Mathematical Terms with Sensor Based Data},
  author={Wegmeth, L. and Hoelzemann, A. and Van Laerhoven, K.},
  journal={arXiv preprint arXiv:2109.05594},
  year={2021}
}

@inproceedings{kress2022hardware,
  title={Hardware-aware Workload Distribution for AI-based Online Handwriting Recognition in a Sensor Pen},
  author={Kre{\ss}, F. and Serdyuk, A. and Hotfilter, T. and others},
  booktitle={2022 11th Mediterranean Conference on Embedded Computing (MECO)},
  pages={1--4},
  year={2022},
  organization={IEEE}
}

@article{klass2022uncertainty,
  title={Uncertainty-aware evaluation of time-series classification for online handwriting recognition with domain shift},
  author={Kla{\ss}, A. and Lorenz, S. M and Lauer-Schmaltz, M. and others},
  journal={arXiv preprint arXiv:2206.08640},
  year={2022}
}

@inproceedings{ott2022domain,
  title={Domain adaptation for time-series classification to mitigate covariate shift},
  author={Ott, F. and R{\"u}gamer, D. and Heublein, L. and others},
  booktitle={Proceedings of the 30th ACM International Conference on Multimedia},
  pages={5934--5943},
  year={2022}
}

@article{azimi2022improving,
  title={Improving Accuracy and Explainability of Online Handwriting Recognition},
  author={Azimi, H. and Chang, S. and Gold, J. and others},
  journal={arXiv preprint arXiv:2209.09102},
  year={2022}
}

@inproceedings{chen2022complex,
  title={Complex Handwriting Trajectory Recovery: Evaluation Metrics and Algorithm},
  author={Chen, Z. and Yang, D. and Liang, J. and others},
  booktitle={Proceedings of the Asian Conference on Computer Vision},
  pages={1060--1076},
  year={2022}
}

@article{nan2021comparison,
  title={Comparison between recurrent networks and temporal convolutional networks approaches for skeleton-based action recognition},
  author={Nan, M. and Tr{\u{a}}sc{\u{a}}u, M. and Florea, A M. and others},
  journal={Sensors},
  volume={21},
  number={6},
  pages={2051},
  year={2021},
  publisher={MDPI}
}

@article{gopali2021comparative,
  title={A Comparative Study of Detecting Anomalies in Time Series Data Using LSTM and TCN Models},
  author={Gopali, S. and Abri, F. and Siami-Namini, S. and others},
  journal={arXiv preprint arXiv:2112.09293},
  year={2021}
}

@inproceedings{huang2022agtgan,
  title={AGTGAN: Unpaired Image Translation for Photographic Ancient Character Generation},
  author={Huang, H. and Yang, D. and Dai, G. and others},
  booktitle={Proceedings of the 30th ACM International Conference on Multimedia},
  pages={5456--5467},
  year={2022}
}

@inproceedings{nguyen2021online,
  title={Online trajectory recovery from offline handwritten Japanese kanji characters of multiple strokes},
  author={Nguyen, H T. and Nakamura, T. and Nguyen, C T. and others},
  booktitle={2020 25th International Conference on Pattern Recognition (ICPR)},
  pages={8320--8327},
  year={2021},
  organization={IEEE}
}

@article{wehbi2022surface,
  title={Surface-Free Multi-Stroke Trajectory Reconstruction and Word Recognition Using an IMU-Enhanced Digital Pen},
  author={Wehbi, M. and Luge, D. and Hamann, T. and others},
  journal={Sensors},
  year={2022},
  publisher={MDPI}
}

@article{krichen2022combination,
  title={Combination of explicit segmentation with Seq2Seq recognition for fine analysis of children handwriting},
  author={Krichen, O. and Corbill{\'e}, S. and Anquetil, {\'E}. and others},
  journal={International Journal on Document Analysis and Recognition (IJDAR)},
  year={2022},
  publisher={Springer}
}

@article{har2002new,
  title={New similarity measures between polylines with applications to morphing and polygon sweeping},
  author={Har-Peled, S. and others},
  journal={Discrete \& Computational Geometry},
  year={2002},
  publisher={Springer}
}

@article{simonnet2019evaluation,
  title={Evaluation of children cursive handwritten words for e-education},
  author={Simonnet, D. and Girard, N. and Anquetil, E. and others},
  journal={Pattern Recognition Letters},
  year={2019},
  publisher={Elsevier}
}

@article{bu2021handwriting,
  title={Handwriting-Assistant: Reconstructing Continuous Strokes with Millimeter-level Accuracy via Attachable Inertial Sensors},
  author={Bu, Y. and Xie, L. and Yin, Y. and others},
  journal={Proceedings of the ACM on Interactive, Mobile, Wearable and Ubiquitous Technologies},
  year={2021},
  publisher={ACM New York, NY, USA}
}

@article{bai2018empirical,
  title={An empirical evaluation of generic convolutional and recurrent networks for sequence modeling},
  author={Bai, S. and Kolter, J Z. and Koltun, V.},
  journal={arXiv},
  year={2018}
}

@inproceedings{guirguis2021seld,
  title={SELD-TCN: Sound event localization \& detection via temporal convolutional networks},
  author={Guirguis, K. and Schorn, C. and Guntoro, A. and others},
  booktitle={2020 28th European Signal Processing Conference (EUSIPCO)},
  year={2021},
  organization={IEEE}
}

@inproceedings{dai2020hybrid,
  title={Hybrid spatio-temporal graph convolutional network: Improving traffic prediction with navigation data},
  author={Dai, R. and Xu, S. and Gu, Q. and others},
  booktitle={Proceedings of the 26th ACM SIGKDD International Conference on Knowledge Discovery \& Data Mining},
  year={2020}
}

@article{yan2020temporal,
  title={Temporal convolutional networks for the advance prediction of ENSO},
  author={Yan, J. and Mu, L. and Wang, L. and others},
  journal={Scientific reports},
  year={2020},
  publisher={Nature Publishing Group}
}

@article{sakoe1978dynamic,
  title={Dynamic programming algorithm optimization for spoken word recognition},
  author={Sakoe, H. and Chiba, S.},
  journal={IEEE transactions on acoustics, speech, and signal processing},
  year={1978},
  publisher={IEEE}
}

@inproceedings{MagHacker,
author = {Liu, Y. and Huang, K. and Song, X. and others},
title = {MagHacker: Eavesdropping on Stylus Pen Writing via Magnetic Sensing from Commodity Mobile Devices},
year = {2020},
isbn = {9781450379540},
publisher = {Association for Computing Machinery},
address = {New York, NY, USA},
url = {https://doi.org/10.1145/3386901.3389030},
doi = {10.1145/3386901.3389030},
booktitle = {Proceedings of the 18th International Conference on Mobile Systems, Applications, and Services},
location = {Toronto, Ontario, Canada},
series = {MobiSys '20}
}

@article{ring,
author = {Wilhelm, M. and Krakowczyk, D. and Albayrak, S.},
year = {2020},
title = {PeriSense: Ring-Based Multi-Finger Gesture Interaction Utilizing Capacitive Proximity Sensing},
journal = {Sensors},
doi = {10.3390/s20143990}
}

@inproceedings{bracelet,
author = {McIntosh, J. and Marzo, A. and Fraser, M.},
title = {SensIR: Detecting Hand Gestures with a Wearable Bracelet Using Infrared Transmission and Reflection},
year = {2017},
isbn = {9781450349819},
address = {New York, NY, USA},
url = {https://doi.org/10.1145/3126594.3126604},
doi = {10.1145/3126594.3126604},
booktitle = {Proceedings of the 30th Annual ACM Symposium on User Interface Software and Technology},
numpages = {5},
location = {Qu\'{e}bec City, QC, Canada},
}

@inproceedings{Shoes,
  TITLE = {{Unsupervised pedestrian trajectory reconstruction from IMU sensors}},
  AUTHOR = {Derrode, S. and Li, H. and Benyoussef, L. and others},
  URL = {https://hal.archives-ouvertes.fr/hal-01786223},
  BOOKTITLE = {{TAIMA 2018: Traitement et Analyse de l'Information M{\'e}thodes et Applications}},
  ADDRESS = {Hammamet, Tunisia},
  YEAR = {2018},
  HAL_ID = {hal-01786223},
}


\section*{Appendix}
\subsection{DTW alignment}
The DTW algorithm \cite{sakoe1978dynamic} is based on dynamic programming to assess the similarity between time series. 
\romain{For two multivariate time series $x \in \mathbb{R}^{T_x \times z}$, and $y \in \mathbb{R}^{T_y \times z}$ of equal feature dimensionality $z$ and respective lengths $T_x$ and $T_y$, the DTW is written as follows:}
\begin{equation}
    DTW(x, y) = \min_{\delta \in E(x, y)} \sum_{(i,j)\in \delta} d(x_i, y_j)
\end{equation}
where $E(x, y)$ represents the set of all admissible alignments between $x$ and $y$ and $d$ is a distance metric in $\mathbb{R}^z$. 
Commonly, the squared Euclidean distance $d(x_i, y_j) =  \|x_i -y_j\| ^2$ is used.

\wassim{An alignment is a sequence of pairs of timestamps that is admissible if (i) it matches the first (and respectively the last) indices of time series x and y together, (ii) it is monotonically increasing, and (iii) it connects the two time series by matching at least one index of each series.}
Using dynamic programming, the admissible alignment paths can be computed according to the following recurrence formula:
\begin{align*}
    DTW(x_{\rightarrow i}, y_{\rightarrow j}) &= d(x_{i}, y_{j}) \\
                                              &+  \min \left\{ \begin{array}{lll}
DTW(x_{\rightarrow i}, y_{\rightarrow j-1}) \\ DTW(x_{\rightarrow i-1}, y_{\rightarrow j}) \\
DTW(x_{\rightarrow i-1}, y_{\rightarrow j-1})
\end{array}\right.
\end{align*}
where $x_{\rightarrow i}$ denotes time series $x$ observed up to timestamp $i$.

\wassim{The DTW method on timestamps links multiple sensor signal points to at least one tablet signal data point.}
In our case, we do not allow the association of multiple points of the tablet signal to one point of the sensor signal. In other words, in the previous recurrence formula the predecessor $DTW(x_{\rightarrow i}, y_{\rightarrow j-1})$ is removed. 
\romain{According to the resulting alignment path, the timestamps from the sensor signal are aligned with their corresponding timestamps from the tablet signal.}

\subsection{Datasets description}
\label{datasets description}
A group of 35 adult writers has contributed to the dataset acquisition. One recording is composed of 34 samples which are randomly selected scripts of characters, words, equations, shapes and word groups. To the next, this new dataset is called IRISA-KIHT. A writer can make several recordings that makes the IRISA-KIHT dataset to be writer imbalanced. 





\begin{table*}
\centering
\footnotesize

\caption{IRISA-KIHT and IRISA-KIHT-S datasets descriptions.}
\label{tbl:ds_dist}
\begin{tabular}{llcccccccc}
Datasets                                                                         & Sets      & \# writers & \# Recordings                                                     & \# Samples & Characters & Words & Equations & Shapes & Word groups  \\ 
\hline
\multirow{3}{*}{\begin{tabular}[c]{@{}l@{}}IRISA-KIHT\end{tabular}} & Training~ & 25            & \begin{tabular}[c]{@{}c@{}}61 (mixed)\\+\\53 (words)\end{tabular} & 3 770      & 915        & 2 306 & 305       & 122    & 122        \\ 
\cline{2-10}
                                                                                 & Test~     & 10             & 10 (mixed)                                                         & 340        & 150         & 100    & 50        & 20      & 20          \\ 
\cline{2-10}
                                                                                 & Total     & 35            & 124~                                                              & 4 110      & 1 065        & 2 406 & 355       & 142    & 142        \\ 
\hline
\multirow{3}{*}{\begin{tabular}[c]{@{}l@{}}IRISA-KIHT-S\\ (3-Folds \\Cross-validation)\end{tabular}} & Training  & 20            & \begin{tabular}[c]{@{}c@{}}20 (mixed)\end{tabular}  & 680        & 300        & 200   & 100       & 40     & 40         \\ 
\cline{2-10}
                                                                                 & Test      & 10             & 10 (mixed)                                                         & 340        & 150         & 100    & 50        & 20      & 20          \\ 
\cline{2-10}
                                                                                 & Total     & 30            & 30~                                                               & 1 020        & 450        & 300   & 150       & 60     & 60         \\
\hline
\end{tabular}
\label{tab:dataset}
\end{table*}

\begin{figure}[ht]
\centering
    \includegraphics[scale=0.3]{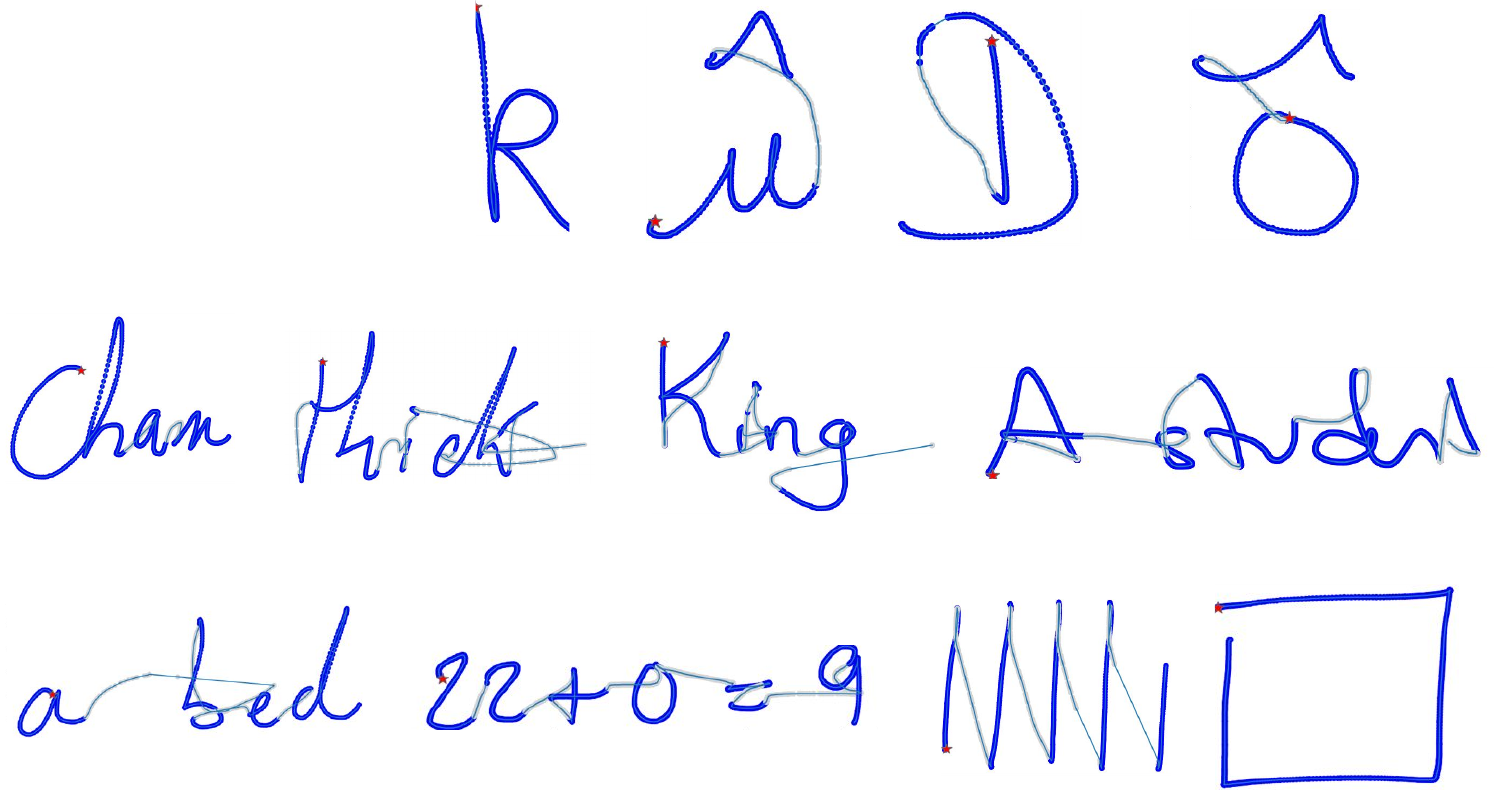}
    \caption{Some examples of IRISA-KIHT data. pen-up (hovering) strokes are in gray and pen-down (touching) strokes are in blue.}
    \label{fig:data}
\end{figure}

To evaluate the generalization capacity of our proposed approach, we decided to train and test our model on different writers. The IRISA-KIHT dataset consists of 25 writers’ recordings in the training set (3,770 samples) which represents around $71.5\%$ of the writers' total number. The rest 10 writers’ recordings never seen in training refer to the test set, representing $28.5\%$ of writers (8.3\% of samples). 
Figure \ref{fig:data} shows some samples of the IRISA-KIHT datasets. 

The IRISA-KIHT-S dataset is a subset of the IRISA-KIHT dataset which is available on request\footnote{Available free of charge for research community upon demand for research purposes only}. 
This dataset is composed of 30 recordings and it is writer balanced as there is one recording per writer. Table \ref{tab:dataset} presents the IRISA-KIHT and IRISA-KIHT-S statistical description.

\wassim{Every 34-sample recording session generates files from the data acquisition mobile app. The sensor signals file has 15 columns and N rows, where N is the number of IMU signals, timestamps, and sensor values. The table has 13 columns: milliseconds, accelerometer front (x, y, z), accelerometer rear  (x, y, z), gyroscope  (x, y, z), magnetometer  (x, y, z), and force signals. Tablet signal files contain milliseconds, position coordinates (x, y, z), and pressure force signals. The transcription (labels) file contains labels and the start and stop time-stamps for every sample. Additional files concerning the sensor calibration and recording meta data are provided. The dataset website describes the format and meta data.}

\subsection{TCN architecture}
A TCN (Temporal Convolutional Network) like layer consists of dilated, residual non-causal 1D convolutional layers with the same input and output lengths. The input tensor of our TCN implementation has the shape (batch\_size=None, input\_length=None, channels\_num=13) and the output tensor has the shape (batch\_size=None, output\_length=None, channels\_num=2). Since each TCN layer has the same input and output length, only the third dimension of the input and output tensors vary ("None" here denote the learning with batches of different size). The TCN layer consists of 4 blocks in TCN-49 and 3 blocks in TCN-373, with each block containing 100 filters followed by a RELU activation function and padded to the "same" input shape. A batch normalization layer was introduced between each pair of consecutive blocks. 
The performance of a TCN architecture based model depends on the size of the receptive field (RF) of the network. Since a TCN’s receptive field depends on the network depth $N$ as well as filter size $k$ and dilation factor $d$, stabilization of deeper and larger TCNs becomes important \cite{bai2018empirical}. The receptive field of our TCN based model is computed as~: 
\begin{equation}
    RF = 1 + 2 \cdot (k - 1) \cdot N \cdot \sum_{i}d_{i}
\end{equation}
The kernel size $k$ is selected so that the receptive field covers enough context for predictions. 
For a regular convolution filter the dilation $d$ is equal to 1. Using larger dilation enables an output at the top level to represent a wider range of inputs.
In the equation, there is a multiplication by 2 because there are two 1D CNN layers in a single residual block of the TCN architecture.
For the purposes of the trajectory reconstruction task, we set the stride equal to 1.
For the TCN-49 we choose to set k=3, N=4 and d={1, 2} and for the TCN-373 k=3, N=3 and d={1, 2, 4, 8, 16}.

\begin{figure}[ht]
\begin{center}
\includegraphics[width=0.4\textwidth]{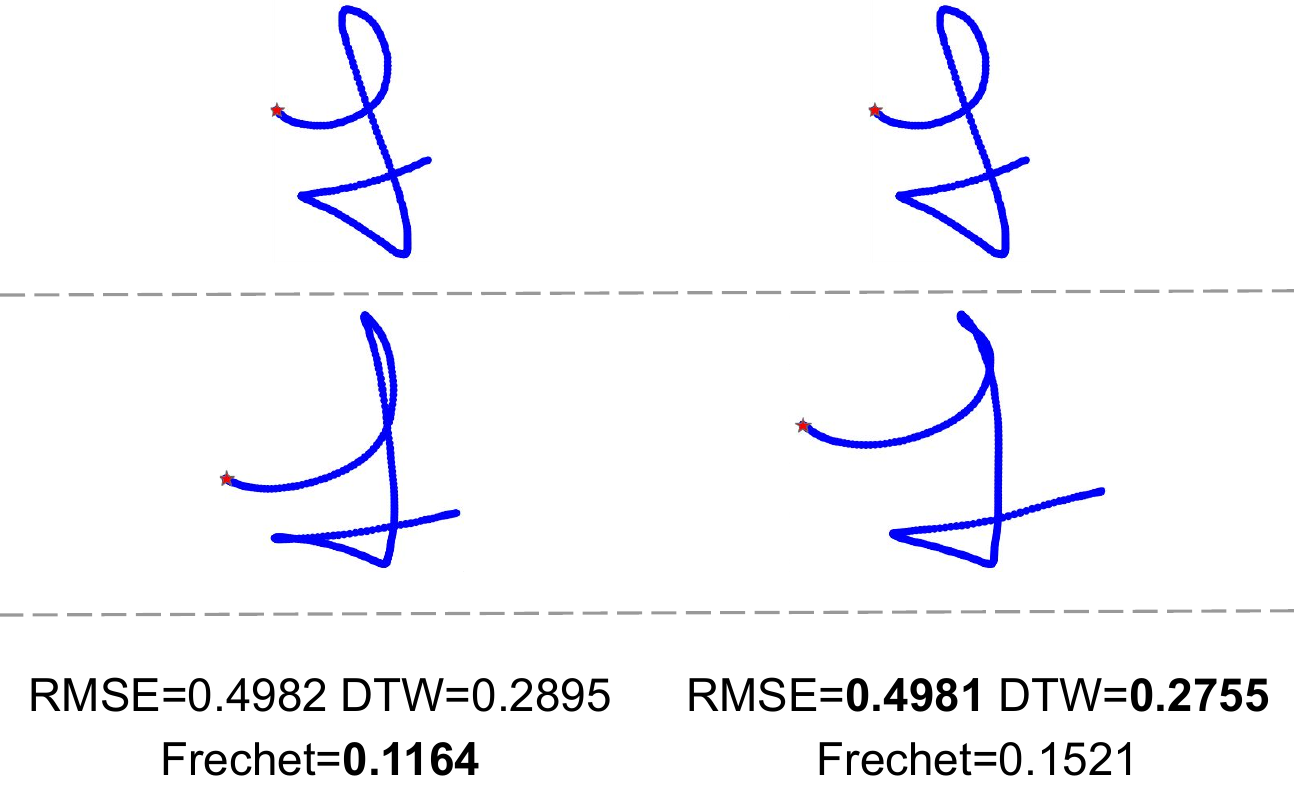}
\end{center}
\caption{Metrics comparison. On the top, the ground truth and on the bottom two predictions coming from different models.}
\label{metric}
\end{figure}

\subsection{Fréchetdistance}
\label{sec:Frechet}
\florent{
The formal definition of the Fréchet distance is: 
Let  $S$ be a metric space,  $d$ its distance function. 
A curve $A$ in $S$ is a continuous map from the unit interval into $S$, i.e. $A:[0,1]\rightarrow S $.
A reparameterization $\alpha$  of $[0,1]$ is a continuous, non-decreasing, surjection $\alpha:[0,1]\rightarrow [0,1]$
Let $A$ and $B$ be two given curves in $S$.
Then, the Fréchet distance between $A$ and $B$ is defined as the infimum over all reparameterizations $\alpha$  and $\beta$  of $[0,1]$ of the maximum over all $t\in [0,1]$ of the distance in $S$ between $A(\alpha (t))$ and $B(\beta (t))$.
The Fréchet distance $F$ is defined by the following equation: 
\begin{equation}
F(A,B)= \inf_{\alpha, \beta} \max_{t\in [0,1]} \left\{ d(A(\alpha(t)), B(\beta(t)) \right\} 
\label{eq:frechet}
\end{equation}}

\vspace{-1cm}

\florent{
Appropriate assessment Metrics are needed to evaluate trajectory reconstruction to give students useful feedback. Fréchet distance doesn't disagree with visual evaluation like DTW and RMSE. Figure \ref{metric} indicates that the left upper loop of the f is better reconstructed. Fréchet distance is needed to observe this.}

\end{document}